\def\ColorScheme{purple}

\documentclass[11pt,logo,letterpaper]{berkeley}

\usepackage[comma,numbers,sort]{natbib}
\usepackage{booktabs}       
\usepackage{nicefrac}       
\usepackage{wrapfig}

\usepackage{multirow}
\usepackage{colortbl}   
\usepackage{tabularx}
\usepackage{float}
\usepackage{placeins}
\usepackage{tcolorbox}
\usepackage{makecell}
\usepackage{xspace}

\usepackage{fontawesome5}
\usepackage{longtable}

\usepackage{tcolorbox}
\tcbuselibrary{breakable, skins} 
\usepackage{listings}
\lstdefinestyle{prompt}{
    breaklines=true,
    breakatwhitespace=false,
    basicstyle=\tiny\ttfamily,
    frame=none,
    columns=flexible,
    breakindent=0pt,
    literate={—}{{-}}1
}
\makeatletter
\newcommand*\myfontsize{%
  \@setfontsize\myfontsize{7}{8}%
}
\makeatother

\definecolor{blanchedalmond}{rgb}{1.0, 0.92, 0.8}
\definecolor{carmine}{rgb}{0.59, 0.0, 0.09}
\definecolor{lightblue}{rgb}{0.22,0.45,0.70}%

\renewcommand{\mathbf}{\boldsymbol}

\makeatletter
\def\Ddots{\mathinner{\mkern1mu\raise\p@
\vbox{\kern7\p@\hbox{.}}\mkern2mu
\raise4\p@\hbox{.}\mkern2mu\raise7\p@\hbox{.}\mkern1mu}}
\makeatother

\definecolor{amaranth}{rgb}{0.9, 0.17, 0.31}
\definecolor{antiquebrass}{rgb}{0.8, 0.58, 0.46}
\definecolor{antiquefuchsia}{rgb}{0.57, 0.36, 0.51}
\definecolor{chromeyellow}{rgb}{0.31, 0.47, 0.26}

\definecolor{guicolor}{RGB}{251, 229, 214}
\definecolor{toolcolor}{RGB}{221, 228, 243}

\definecolor{deepgreen}{RGB}{0, 100, 0}
\definecolor{deepred}{RGB}{139, 0, 0}

\definecolor{backblue}{RGB}{210, 230, 250}

\definecolor{backblue}{RGB}{210, 230, 250}
\definecolor{backgreen}{RGB}{226, 240, 217}
\definecolor{backred}{RGB}{255, 223, 223}
\definecolor{mycitecolor}{HTML}{1B0EAA}

\definecolor{citecolor}{RGB}{0,255,0}
\definecolor{refcolor}{RGB}{255,0,0}
\definecolor{urlcolor}{RGB}{0,180,255}

\definecolor{my_blue}{RGB}{225, 234, 255}
\definecolor{my_red}{RGB}{253, 241, 240}
\definecolor{my_green}{RGB}{239, 250, 238}

\definecolor{my_green2}{HTML}{e6ffe6}
\definecolor{my_blue2}{HTML}{d4f3fd}
\definecolor{my_purple}{HTML}{8b71d7}
\definecolor{my_purple2}{HTML}{dccbe6}

\definecolor{deepgreen}{RGB}{0, 80, 40} 
\definecolor{lightgreen}{RGB}{235, 245, 235}

\definecolor{dropred}{RGB}{255, 230, 230} 
\definecolor{gainpurple}{RGB}{235, 230, 250} 
\definecolor{textred}{RGB}{200, 0, 0}
\definecolor{textblue}{RGB}{0, 0, 150}

\title{ToolCUA: Towards Optimal GUI-Tool Path Orchestration for Computer Use Agents}
\runningtitle{ToolCUA: Towards Optimal GUI-Tool Path Orchestration for Computer Use Agents}
\author{
    Xuhao, Hu$^{2,1,*}$,
    Xi Zhang$^{1,*}$,
    Haiyang Xu$^{1,\dagger}$,
    Kyle Qiao$^{1}$,
    Jingyi Yang$^{2}$ \\
    Xuanjing Huang$^{2}$,
    Jing Shao$^{3}$,
    Ming Yan$^{1,\dagger}$,
    Jieping Ye$^{1}$
}
\affil{$^1$Tongyi Lab, Alibaba Group \quad $^2$Fudan University \quad $^3$Shanghai Artificial Intelligence Laboratory}

\correspondingauthor{$^*$Equal Contribution, $\dagger$Corresponding Author}

\begin{document}

\begin{abstract}

Computer Use Agents (CUAs) can act through both atomic GUI actions (\textit{e.g.}, clicking and typing) and high-level tool calls (\textit{e.g.}, API-based file operations), but they are often confused by this hybrid action space: they fail to determine when to continue with GUI actions and when to switch to tools, leading to suboptimal execution paths.
%
%
This difficulty stems from two issues.
First, high-quality interleaved GUI-Tool trajectories are scarce, and collecting real tool trajectories is expensive and brittle.
Second, existing supervision provides limited guidance for GUI-Tool path selection, as most methods focus on step-level action imitation or final task completion and offer little trajectory-level feedback on whether GUI-Tool switching leads to a more effective execution path.
%
In this paper, we propose \textbf{ToolCUA},
an end-to-end agent designed to learn optimal GUI-Tool path selection through a staged training paradigm.
%
We first introduce an \textbf{Interleaved GUI-Tool Trajectory Scaling Pipeline} that repurposes abundant static GUI trajectories and synthesizes a grounded library of simulated tools, making it possible to scale diverse GUI-Tool trajectories without manual engineering or real tool-trajectory collection.
%
%
Based on this data, we perform Tool-Bootstrapped GUI RFT, which combines warmup SFT with single-turn RL to improve decisions at critical GUI-Tool switching points.
%
Finally, we further optimize ToolCUA with \textbf{Online Agentic RL} in a high-fidelity GUI-Tool environment, using a \textbf{Tool-Efficient Path Reward} that encourages both appropriate tool use and shorter execution paths.
%
Experiments on OSWorld-MCP show that ToolCUA achieves 46.85\% accuracy, a relative improvement of approximately 66\% over the baseline, establishing a new state of the art among models of comparable scale.
It also improves by 3.9 percentage points over the GUI-only setting, demonstrating effective GUI-Tool orchestration.
The results further suggest that training in a hybrid action space is a promising paradigm for real-world digital agents.

\vspace{0.5cm}
\textbf{Date}: May 12, 2026

\textbf{Author emails}: \email{\{xuhaohu08\}@gmail.com}, \email{\{zx443053\}@alibaba-inc.com}

\textbf{Correspondence}: \email{\{shuofeng.xhy,ym119608\}@alibaba-inc.com}

\textbf{\faGithub~~Code}: \url{https://github.com/X-PLUG/ToolCUA}
\end{abstract}

\maketitle

\begin{figure}[H]
  \vspace{-2mm}
  \centering
  \includegraphics[width=0.98\textwidth]{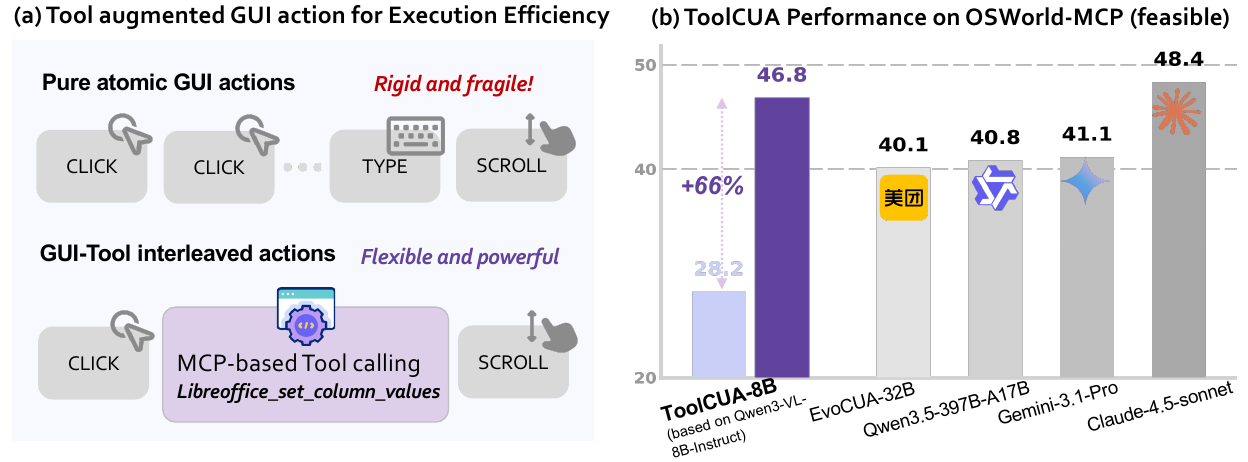}
  \vspace{-2mm}
  \caption{(a) The advantage of Tool-augmented actions compared with pure GUI actions. (b) The performance of our \textbf{ToolCUA} compared with the baselines, agentic CUAs, and general models.}
  \label{fig:main_teaser}
  \vspace{-6mm}
\end{figure}

\section{Introduction}
\label{sec:intro}

The rapid evolution of Multimodal Large Language Models (MLLMs)~\cite{anthropic2025claudeopus45,bai2025qwen3,bai2025qwen2,chen2024internvl,zeng2026glm5, team2026kimi} toward agentic capabilities~\cite{li2025mm,wei2026agenticmme,ye2026claw,wang2026openclawrl} has established Computer Use Agents (CUAs)~\cite{openai2025operator, wang2024mobile, wang2025opencua, xu2026mobilev35, qin2025ui, wang2025ui, liu2025scalecua, yan2025step, zhang2025ufo} as a frontier topic for automating native desktop workflows.
%
%
Conventionally, CUAs primarily rely on atomic GUI actions (\textit{e.g.}, clicking and scrolling),
which offer broad generalizability but are susceptible to cascading errors in long-horizon tasks.
%
%
In contrast, structured tool calls~\cite{team2025tongyi, qin2023toolllm, feng2025retool, wei2026agenticmme} provide agents with superior efficiency and precision~\cite{zhang2025apiagent, zhang2025ufo}.
%
For example, in Figure~\ref{fig:main_teaser} (a),
modifying an entire column in LibreOffice can be completed by a single API call, whereas a pure GUI solution requires a long sequence of clicks and keystrokes.
However, tool-based APIs are constrained by service coverage and stability, limiting applicability in diverse scenarios.
Therefore, given their complementary strengths, a hybrid GUI-Tool action space is essential for next-generation CUAs.

\begin{figure}[!b]
\vspace{-4mm}
\centering
\begin{minipage}[t]{0.61\textwidth}
\vspace{0pt}
\centering
\small
\captionof{table}{Performance comparison between pure GUI and hybrid GUI-Tool action spaces on the matched feasible task set of OSWorld~\cite{xie2024osworld} and OSWorld-MCP~\cite{jia2025osworldmcp}. ``Steps'' is the average number of trajectory steps. For the hybrid setting, ``Tool-calls'' is averaged over successful trajectories in the Tool-Beneficial subset. See details in Appendix~\ref{sec:appendix_preliminary_study}.}
\label{tab:mcptool_coordination_failure}
\resizebox{1.0\linewidth}{!}{\begin{tabular}{llccc}
\toprule
\textbf{Model} & \textbf{Action} & \textbf{Accuracy $\uparrow$} & \textbf{Steps $\downarrow$} & \textbf{Tool-calls} \\
\midrule

\multirow{2}{*}{Qwen3VL-8B} & GUI & 29.0 &  19.2 & - \\
& + Tools & \cellcolor{dropred}28.2 {\color{textred}\scriptsize(\textbf{$-$0.8})} & \cellcolor{dropred}19.3 & \cellcolor{gainpurple}0.01 \\
\addlinespace[0.5em]

\multirow{2}{*}{Qwen3VL-235B} & GUI & 41.1 & 25.9 & - \\
& + Tools & \cellcolor{dropred}38.1 {\color{textred}\scriptsize(\textbf{$-$3.0})} & \cellcolor{dropred}18.0 & \cellcolor{gainpurple}{{3.62}} \\
\addlinespace[0.5em]

\multirow{2}{*}{EvoCUA-32B} & GUI & 52.6 & 25.0 & - \\
& + Tools & \cellcolor{dropred}40.5 {\color{textred}\scriptsize(\textbf{$-$12.1})} & \cellcolor{dropred}26.2 & \cellcolor{gainpurple}{{6.23}} \\
\addlinespace[0.5em]

\multirow{2}{*}{Claude-4-Sonnet} & GUI & 47.7 & 23.6 & - \\
& + Tools & \cellcolor{dropred}43.5 {\color{textred}\scriptsize(\textbf{$-$4.2})} & \cellcolor{dropred}19.8 & \cellcolor{gainpurple}4.30 \\
\addlinespace[0.5em]

\multirow{2}{*}{Claude-4.5-Sonnet} & GUI & 61.9 & 23.3 & - \\
& + Tools & \cellcolor{dropred}48.4 {\color{textred}\scriptsize(\textbf{$-$13.5})} & \cellcolor{dropred}19.1 & \cellcolor{gainpurple}{{3.81}} \\

\midrule
\multirow{2}{*}{\textbf{ToolCUA-8B (Ours)}} & GUI & 42.9 & 19.4 & - \\
& + Tools & \cellcolor{lightgreen}\textbf{46.8} {\color{deepgreen}\scriptsize(\textbf{$+$3.9})} & \cellcolor{lightgreen}\textbf{14.9} & \cellcolor{gainpurple}\textbf{2.11} \\
\bottomrule
\end{tabular}}
\end{minipage}
\hfill
\begin{minipage}[t]{0.37\textwidth}
\vspace{-1pt}
\centering
\includegraphics[width=1.0\textwidth]{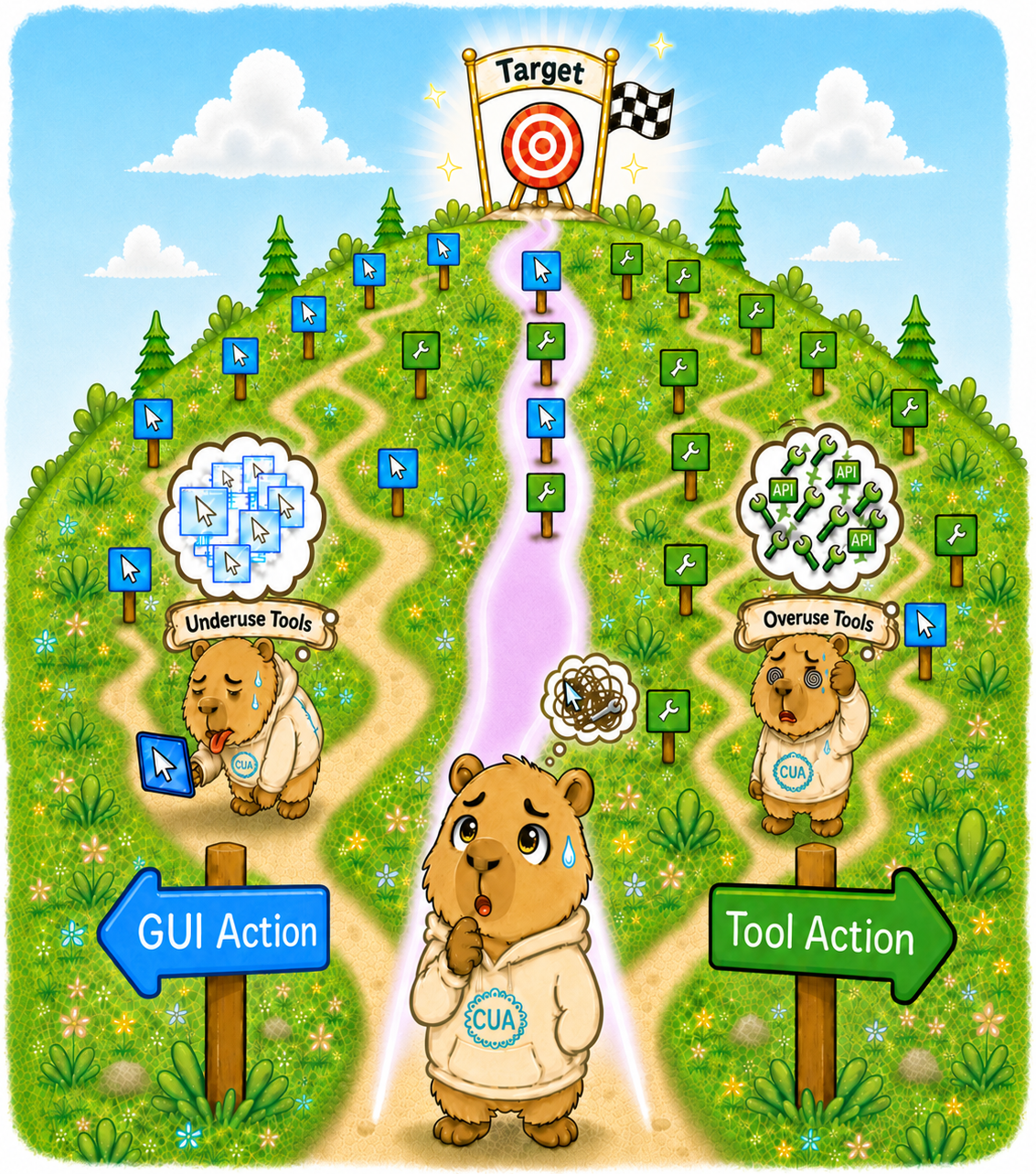}
\vspace{-5mm}
\captionof{figure}{Current computer use agents suffer from optimal path confusion under GUI-Tool hybrid actions.}
\label{fig:cartoon_fig}
\end{minipage}
\end{figure}

Although GUI actions and tool calls are complementary, simply exposing both action spaces to an MLLM does not solve the problem. In practice, agents are often \textbf{confused by the hybrid action space}. As shown in Table~\ref{tab:mcptool_coordination_failure}, some models invoke tools frequently on successful tool-beneficial tasks yet suffer lower overall task success,
while others (\textit{e.g.}, Qwen3VL-8B) remain overly GUI-centric and almost never invoke tools even when more efficient tool calls are available.
%
%
We formalize this challenge, illustrated in Figure~\ref{fig:cartoon_fig}, as \textbf{optimal GUI-Tool path selection}:
dynamically determining when to use GUI actions and when to invoke tools so as to form an efficient and reliable task trajectory.
%
%
%
Unlike step-level action selection, this is inherently a trajectory-level policy learning problem, as each GUI-to-Tool or Tool-to-GUI switching decision not only solves the immediate step, but reshapes the entire subsequent trajectory in terms of efficiency and reliability.
%
%

%
To this end, existing approaches fall short in two fundamental aspects.
First, \textbf{current CUAs are often undertrained on tool use, exhibiting a deficit in tool-calling knowledge}.
This limitation is rooted in the lack of high-quality interleaved GUI-Tool trajectories. In real computer-use environments, usable tools are difficult to obtain and maintain.
Specifically, APIs are often application-specific, incomplete, or unstable, and collecting GUI-Tool data requires expensive environment instrumentation.
Existing efforts~\cite{yang2025ultracua, yan2025step} partly address this by generating tools from code, but such pipelines remain costly to scale and do not fully exploit the large amount of existing GUI-only trajectory corpora~\cite{jian2026cuasuit, wang2025opencua, liu2025scalecua, xie2024osworld, mu2025gui360, zhang2026tongui}.
%
%
Second, even when basic tool-use ability is available, existing supervision provides \textbf{limited guidance for learning effective hybrid action orchestration}.
%
In practice, current training signals usually come from either step-level imitation or final task-completion rewards.
The former captures only local action plausibility, while the latter does not distinguish between a timely tool-call switch and a long, brittle GUI-only workaround.
As a result, the model cannot reliably learn whether switching between GUI actions and tool calls improves the full trajectory.


%


To address these challenges, we introduce \textbf{ToolCUA}, a unified agentic model trained through a staged paradigm. It first builds hybrid-action foundations with scalable interleaved GUI-Tool data and then improves trajectory-level GUI-Tool decisions through reinforcement learning.
%
First, we propose an \textbf{interleaved GUI-Tool trajectory scaling pipeline} built on existing static GUI corpora.
It employs MLLMs to synthesize a trajectory-aware library of tools from recurrent GUI procedures and converts GUI-only data into interleaved GUI-Tool trajectories through tool-step generation and next-state grounding.
By repurposing existing GUI corpora and synthesizing tools instead of collecting expensive real tool trajectories, this pipeline enables scalable data construction without manual engineering, while covering varied tool granularities and switching contexts.
%
%
%
%
Building on this data, we perform Tool-Bootstrapped GUI Reinforcement Fine-Tuning (RFT). It first applies warmup SFT to establish basic hybrid-action capabilities, and then uses single-turn RL to improve decisions at explicit GUI-Tool switching points.
%
%
Finally, we optimize ToolCUA with \textbf{Online Agentic RL} in a realistic GUI-Tool environment with a \textbf{Tool-Efficient Path Reward}, which includes a tool appropriateness term $R_{tool}$ and path efficiency term $R_{length}$:
$R_{tool}$ incentivizes the agent to invoke tools when beneficial and abstain when unnecessary, while $R_{length}$ encourages shorter execution paths by replacing redundant GUI actions with tool calls.
Together, they provide trajectory-level feedback that drives the model toward globally optimal GUI-Tool path selection.

%
Experimental results demonstrate that ToolCUA achieves 46.85\% accuracy on OSWorld-MCP~\cite{jia2025osworldmcp}, establishing state-of-the-art performance among models of comparable scale. This represents an approximately 66\% relative improvement over the Qwen3-VL-8B-Instruct baseline~\cite{bai2025qwen3} and is competitive with leading proprietary models~\cite{anthropic2025claudeopus45, deepmind2026gemini31}.
Furthermore, ToolCUA achieves 42.9\% accuracy in the pure-GUI setting and gains another 3.9 percentage points when tools are available, demonstrating effective GUI-Tool path selection.
Additionally, ToolCUA shows out-of-distribution generalization across tasks and platforms, reaching 23.9\% on unseen \textit{multi\_apps} Linux tasks and achieving 33.8\% on unseen \textit{Windows} desktop apps in WindowsAgentArena~\cite{bonatti2024windows}.
These results confirm that operating in a hybrid GUI-Tool action space is essential for achieving generalizable and efficient real-world digital automation.
%
%
Our main contributions are summarized as follows:
\begin{itemize}[left=3mm, itemsep=0pt, topsep=0pt, parsep=2pt, partopsep=0pt]

\item We propose an \textbf{Interleaved GUI-Tool trajectory scaling pipeline} that repurposes existing pure GUI corpora into scalable hybrid-action training data through tool synthesis, obviating the need for manual tool-environment construction and tool trajectory collection.


\item We propose a staged training paradigm consisting of \textbf{Tool-Bootstrapped GUI RFT}, which establishes hybrid-action foundations and calibrates GUI-Tool switching decisions, followed by \textbf{Online Agentic RL} with a \textbf{Tool-Efficient Path Reward} ($R_{tool}$ and $R_{length}$) for trajectory-level optimization.

\item Our \textbf{ToolCUA} reaches 46.85\% accuracy on OSWorld-MCP, achieving state-of-the-art performance among similarly sized models and outperforming its pure-GUI counterpart. Our findings suggest that training in a hybrid GUI-Tool action space enables more generalizable and efficient computer-use automation.

\end{itemize}

\FloatBarrier

\section{ToolCUA}
\label{sec:method}

\subsection{Definition and Scope}

We first formalize the computer-use task as a Markov Decision Process (MDP) where $\mathcal{M} = \langle \mathcal{S}, \mathcal{A}, \mathcal{P}, \mathcal{R}, \gamma \rangle$.
At each time step $t$, the state $s_t \in \mathcal{S}$ denotes a multimodal observation encompassing both the desktop screenshot and previously invoked tool results.
The agent interacts with the environment through a hybrid action space $\mathcal{A} = \mathcal{A}_{GUI} \cup \mathcal{A}_{Tool}$, where $a_t \in \mathcal{A}_{GUI}$ represents atomic GUI interactions such as coordinate-based clicks, and $a_t \in \mathcal{A}_{Tool}$ signifies high-level structured tool invocations.
The objective is to learn an optimal policy $\pi_\theta(a_t | s_t)$ that maximizes the expected cumulative reward over a long-horizon trajectory $\tau$:
\begin{equation}
    J(\theta) = \mathbb{E}_{\pi_\theta} \left[ \sum_{t=0}^{T} \gamma^t \mathcal{R}(s_t, a_t) \right]
\end{equation}



%

%

%
\subsection{Interleaved GUI-Tool Trajectory Scaling Pipeline}
\label{sec:scaling_pipeline}


%

\begin{figure}[t]
  \centering
  \small
  \includegraphics[width=1.0\textwidth]{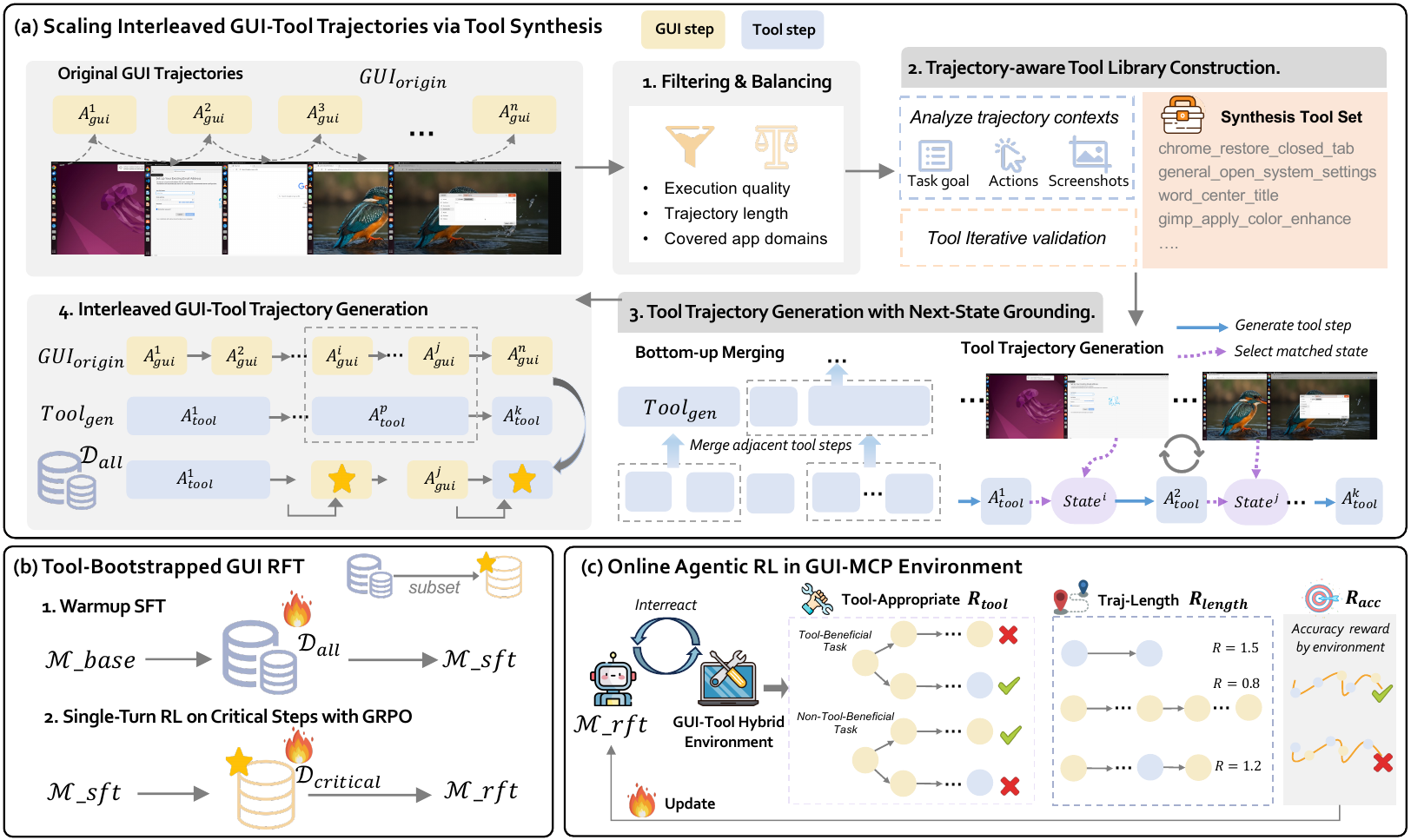}
  \vspace{-2mm}
  \caption{
   Overview of the ToolCUA data collection and training paradigm. (a)
   Interleaved GUI-Tool trajectory scaling via offline GUI data and tool synthesis, yielding the dataset $\mathcal{D}_{all}$ containing critical switching steps $\mathcal{D}_{critical}$.
   (b) Tool-Bootstrapped GUI RFT with warmup SFT on $\mathcal{D}_{all}$  and step-level RL on
   $\mathcal{D}_{critical}$. (c) Online Agentic RL with Tool-Efficient Path Reward ($R_{tool}$ and $R_{length}$) for trajectory-level optimization with appropriate tool use and efficient path selection.
}
\label{fig:method_overview}
  \vspace{-2mm}
\end{figure}

To address the scarcity of interleaved GUI-Tool trajectories,
we build an offline trajectory scaling pipeline that starts from \textbf{existing pure GUI trajectories} and converts them into interleaved GUI-Tool data.
As shown in Figure~\ref{fig:method_overview}(a), the key idea is to use an MLLM (\textit{e.g.}, Kimi-K2.5 or Claude-4.5-Sonnet) to synthesize a \textbf{grounded library of simulated tools} from recurrent GUI procedures, and then use these tools to transform GUI-only trajectories into interleaved GUI-Tool trajectories. These simulated tools are symbolic definitions and calls used for execution-free offline supervision; they do not require concrete tool implementation or sandbox execution at this stage.
Our pipeline scales data along three dimensions: {Tool Functionality} across application domains, {Tool Granularity} from atomic utilities to composite skills, and {GUI-Tool Switching Context} covering cases where tool use is more or less beneficial.
Please refer to Appendix~\ref{sec:appendix_synthesis_pipeline_prompt} for prompts we used.
We describe the main steps below.

\noindent\textbf{Trajectory Filtering and Balancing.} We start from successful raw GUI trajectories and filter them by execution quality, task length, and application coverage. %
The remaining trajectories are balanced across domains to provide a stable source distribution for tool synthesis.

\noindent\textbf{{Trajectory-Aware Synthetic Tool Library Construction.}}
For each GUI trajectory, we use an MLLM to synthesize a candidate library of tools by analyzing the {pure GUI path}, including the task goal, action sequences, and dense screenshot descriptions.
Each synthesized tool abstracts an observed GUI procedure into a callable high-level operation, specified by a functional signature, natural language description, and argument semantics inferred from the trajectory.
This makes the tools grounded in concrete trajectory behavior rather than generic API templates or manually predefined functions.
To increase \textbf{diversity}, we synthesize tools at
varying levels of specificity, from single-action wrappers (\textit{e.g.}, \texttt{chrome\_open\_settings}) to multi-step composite functions (\textit{e.g.}, \texttt{chrome\_open\_language\_settings}).
A rule-based format verification is also applied for tool filtering.


\noindent\textbf{Tool Trajectory Generation with Next-State Grounding.} Given the synthesized tool library and the original GUI trajectory, we adopt an MLLM to generate a functionally equivalent tool-only trajectory.
For each step, the MLLM selects an appropriate tool from the library, produces a chain-of-thought rationale, and predicts the expected response; the generated call is then validated against the tool schema.
%
Furthermore, we use an MLLM to perform \textbf{next-state grounding}, \textit{i.e.}, to anchor each simulated tool call to the real next-state screenshot reached by the corresponding contiguous GUI segment in the original successful trajectory. This verifies that the predicted tool effect is consistent with the observed GUI state while preserving the verified endpoint of the source demonstration. Candidates that fail schema validation or next-state grounding are discarded.
We also apply a \textbf{bottom-up merging strategy}: adjacent fine-grained steps sharing a common sub-goal are progressively merged into higher-level composite tool calls, yielding multiple variants at different levels of tool granularity.

\begin{figure}[t]
  \centering
  \small
  \includegraphics[width=1.0\textwidth]{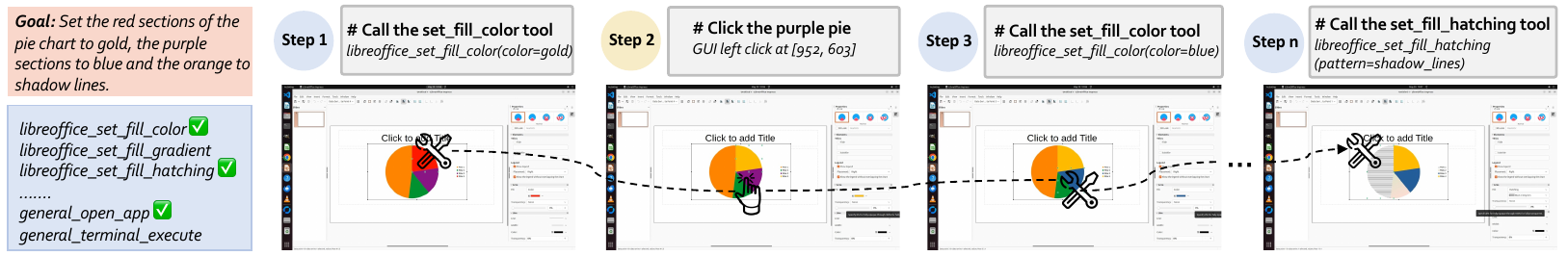}
  \vspace{-2mm}
 \caption{A synthetic interleaved GUI-Tool trajectory generated by our pipeline, demonstrating strategic tool selection and seamless switching between atomic GUI actions and tool calls.}
\label{fig:synthesis_data_example}
  \vspace{-4mm}
\end{figure}

\noindent\textbf{Interleaved GUI-Tool Trajectory Generation.}
Given a grounded tool-only trajectory, we randomly sample a subset of tool calls and replace each with its corresponding GUI action sequence from the original trajectory.
Notably, the replaced tools are simultaneously removed from the tool library, constructing a partial tool-availability context where the agent must fall back to GUI operations when certain tools are unavailable.
By varying the selection of replaced tool calls, we generate diverse interleaved variants from the same trajectory, which are aggregated into $\mathcal{D}_{all}$.
A representative interleaved GUI-Tool trajectory is illustrated in Figure~\ref{fig:synthesis_data_example}.
Furthermore, each replacement naturally exposes two types of boundary transitions: GUI $\rightarrow$ Tool and Tool $\rightarrow$ GUI (\textit{i.e.}, the yellow star in Figure~\ref{fig:method_overview}(a)), where the agent switches between GUI and tool calls.
We refer to these as \textbf{critical switching steps} and collect them into $\mathcal{D}_{critical}$.

\subsection{Tool-Bootstrapped GUI RFT}


With $\mathcal{D}_{all}$ and $\mathcal{D}_{critical}$, we perform Tool-Bootstrapped GUI RFT to equip the baseline agent $\mathcal{M}_{base}$ with flexible hybrid-action behavior and calibrate local GUI-Tool decisions.

\textbf{Warmup Supervised Fine-Tuning (SFT).} %
We first perform SFT on $\mathcal{D}_{all}$ using a standard cross-entropy loss $\mathcal{L}_{SFT} = -\sum \log \pi_\theta(a_t | s_t)$. %
This phase teaches the model multimodal tool-calling knowledge in the CUA domain, including tool selection, argument generation, and the resulting states after tool execution. This SFT warmup yields the model $\mathcal{M}_{sft}$.

\textbf{Single-Turn RL on Critical Steps.} %
Building upon the model $\mathcal{M}_{sft}$, we implement a single-turn RL phase using the Group Relative Policy Optimization (GRPO) algorithm~\cite{shao2024deepseekmath} on $\mathcal{D}_{critical}$. %
By sampling multiple completions at these critical switching steps, the model receives direct feedback on whether to continue with GUI actions or switch to tool calls when appropriate tools are available.
This targeted optimization calibrates the model's discernment at decision boundaries, yielding a coordinated agent $\mathcal{M}_{rft}$ ready for long-horizon online exploration in the GUI-Tool environment. %

%

\subsection{Online Agentic RL with Tool-Efficient Path Reward in GUI-Tool Environment}

%

Online RL extends step-level tool-calling knowledge to complete trajectories, enabling the agent to discover which GUI-Tool switching strategies lead to successful outcomes through real environment exploration.
However, task success alone cannot distinguish whether tool usage was genuinely appropriate, nor whether the execution path was unnecessarily long.
%
%
%
Therefore, we introduce a \textbf{Tool-Efficient Path Reward} that explicitly shapes the agent toward tool-appropriate and efficient trajectories,
which consists of a tool appropriateness term $R_{tool}$ and path efficiency term $R_{length}$:
%
\begin{equation}
\label{eq:reward_shape}
\begin{gathered}
    R = R_{fmt} + R_{acc} + \lambda \cdot R_{tool} + \beta \cdot R_{length} \\[5pt]
\end{gathered}
\end{equation}
where $R_{fmt}$ and $R_{acc}$ are standard format and accuracy rewards, while $R_{tool}$ and $R_{length}$ are activated only when the trajectory succeeds.


\noindent\textbf{Tool Appropriateness Reward Term.}
In practice, agents may complete a task without tools even when tools would help, or invoke tools unnecessarily on tasks that do not require them.
$R_{tool}$ addresses this by introducing a task-level tool-beneficial label $t_b \in \{1, -1\}$ annotated during data construction.
%
Here, $t_b=1$ indicates that the task favors tool usage and $t_b=-1$ indicates that tool usage is unnecessary. 
Let $c$ denote the cumulative number of tool calls in a trajectory.
\begin{equation}
\label{eq:reward_tool}
\begin{gathered}
    R_{tool} = \mathbb{I}_{succ} \cdot \mathbb{I}[(t_b > 0 \land c > 0) \lor (t_b < 0 \land c = 0)] \\[2pt]
\end{gathered}
\end{equation}
$R_{tool}$ is assigned when the agent invokes tools on tool-beneficial tasks ($t_b=1, c>0$), or when it deliberately abstains from tools on non-tool-beneficial tasks ($t_b=-1, c=0$).
This design augments task success with an explicit tool-appropriateness signal, encouraging the agent to use tools when and only when they are beneficial.

\noindent\textbf{Path Efficiency Reward Term.}
Even when the agent succeeds and uses tools appropriately, it may still take unnecessarily long paths, for example, by relying on redundant GUI operations when a single tool call could accomplish the same effect.
To this end,
$R_{length}$ encourages the agent to actively explore and discover more efficient GUI-Tool execution paths through online interaction.
%
Rather than measuring efficiency against a fixed threshold, we evaluate trajectory length relative to the rollout group,
where $s$ is the current trajectory's step count, $\bar{s}$ is the group-average step count, and $S_{max}$ is the maximum execution horizon.

\begin{equation}
\label{eq:reward_length}
\begin{gathered}
    R_{length} = \mathbb{I}_{succ} \cdot \begin{cases} (1 + \frac{\bar{s} - s}{\bar{s}}) & s < \bar{s} \\ \exp(-\frac{s - \bar{s}}{S_{max} - \bar{s}}) & s \geq \bar{s} \end{cases}
\end{gathered}
\end{equation}
For trajectories shorter than the group average, the agent receives a linear bonus proportional to the relative step reduction; otherwise, the reward decays exponentially as the trajectory grows longer.
Since useful tool calls often replace multiple atomic GUI operations, this signal naturally incentivizes the agent to switch to tools when they lead to a shorter and more reliable execution path.

With the above reward, we optimize ToolCUA using multi-turn GRPO over online rollouts in a GUI-Tool environment.
Inspired by DAPO~\cite{yu2025dapo}, we apply dynamic filtering and retain only rollout groups containing both successful and failed trajectories, which improves the informativeness of group-relative policy updates while reducing unnecessary computation.

%

\section{Experiments}
\label{sec:experiments}

\subsection{Experimental Settings}
\label{sec:exp_settings}

\noindent\textbf{Implementation Details.} Our pipeline aggregates diverse trajectories from open-source datasets~\cite{wang2025opencua,liu2025scalecua}, as detailed in Appendix~\ref{sec:appendix_imple_details_data}.
%
We adopt Qwen3-VL-8B-Instruct~\cite{bai2025qwen3} as our base model $\mathcal{M}$.
In the warmup SFT stage, we train $\mathcal{M}$ for 3 epochs and subsequently conduct single-turn RL with a group size of 32.
During the subsequent online agentic RL stage, we set the reward weights to $\lambda = 0.4$ and $\beta = 0.2$, and set $S_{max} = 30$ as the maximum execution horizon.
The training configuration for this stage includes a rollout size of 32 per group, a learning rate of $1 \times 10^{-6}$, and a training batch size of 32 to obtain \textbf{ToolCUA}.
%
%
We further optimize the tool-calling interface by designing an agent-readable return format that provides concise, semantically dense feedback to reduce token overhead and improve grounding accuracy. %
For online agentic RL, we use a mixture of original, rewritten, and newly scaled single-domain tasks, while excluding the entire \textit{multi\_apps} domain for held-out evaluation. Please refer to Appendix~\ref{sec:appendix_imple_details_training} for the task composition and construction details.

%

\noindent\textbf{Baselines and Benchmark.} We evaluate ToolCUA against two categories: general-purpose foundation models, including Qwen3.5-Plus~\cite{qwen3.5}, Claude-4.5-Sonnet~\cite{anthropic2025claudeopus45}, and Gemini-3.1-Pro~\cite{deepmind2026gemini31}; and specialized CUAs, including UI-Tars-1.5~\cite{qin2025ui}, the EvoCUA series~\cite{xue2026evocua}, and GUI-Owl-1.5~\cite{xu2026mobilev35}.

For evaluation, we use OSWorld-MCP~\cite{jia2025osworldmcp} as our primary benchmark. It evaluates CUAs in a hybrid action space covering standard GUI actions, more than 150 tools, and mainstream desktop applications.
%
Following the benchmark setup, we report results on the feasible tasks only.
To mitigate environmental stochasticity in the sandbox, we report \textit{average@3} for all primary metrics and set the maximum number of steps per task to $50$.
We follow the original benchmark metrics (detailed in Appendix~\ref{sec:appendix_imple_details_evaluation}), where TIR measures whether the agent uses tools when beneficial and avoids them when unnecessary, and ACS measures average completion steps as an indicator of execution efficiency.
Furthermore, we evaluate the cross-task and cross-platform transferability of ToolCUA on unseen Linux \textit{multi\_apps} tasks and unseen Windows apps in WindowsAgentArena~\cite{bonatti2024windows}.
%

\begin{table}[!t]
\centering
\small
\caption{Performance on OSWorld-MCP. Results are reported on the feasible tasks only. Tool-Beneficial Tasks favor tool usage, while Non-Tool-Beneficial Tasks do not require tools. Accuracy denotes average success rate; TIR (Tool Invocation Rate) measures whether the agent uses tools when beneficial and avoids them when unnecessary; ACS denotes Average Completion Steps.}
\resizebox{1.0\textwidth}{!}{
\begin{tabular}{l ccc ccc ccc}
\toprule
\multirow{2}{*}{\textbf{Agent Model}} &
\multicolumn{3}{c}{\textbf{Tool-Beneficial Tasks (238)}} &
\multicolumn{3}{c}{\textbf{Non-Tool-Beneficial Tasks (95)}} &
\multicolumn{3}{c}{\textbf{Overall (333)}} \\
\cmidrule(lr){2-4} \cmidrule(lr){5-7} \cmidrule(lr){8-10}
& \textbf{Accuracy} & \textbf{TIR} & \textbf{ACS}
& \textbf{Accuracy} & \textbf{TIR} & \textbf{ACS}
& \textbf{Accuracy $\uparrow$} & \textbf{TIR $\uparrow$} & \textbf{ACS $\downarrow$} \\
\midrule
\multicolumn{10}{c}{\textit{General Model}} \\
\midrule
Gemini-2.5-Pro       & 24.79 & 21.15 & 26.12 & 8.77  & 7.37  & 39.60 & 20.22 & 17.22 & 29.97 \\
OpenAI o3            & 26.89 & 24.51 & 26.71 & 4.91  & 2.46  & 44.78 & 20.62 & 18.22 & 31.87 \\
Seed1.5-VL           & 34.03 & 23.81 & 18.99 & 35.79 & 34.39 & 24.93 & 34.53 & 26.83 & 20.69 \\
Claude-4-Sonnet      & 45.24 & 36.13 & 18.29 & 39.30 & 34.74 & 21.09 & 43.54 & 35.74 & 19.76 \\
Gemini-3.1-Pro       & 44.54 & 37.39 & 22.46 & 32.63 & 26.32 & 32.82 & 41.14 & 34.23 & 25.40 \\
Claude-4.5-Sonnet    & 50.00 & 42.02 & 17.93 & 44.21 & 35.79 & 21.91 & 48.35 & 40.24 & 19.07 \\
Qwen3-VL-235B-A22B   & 37.11 & 28.15 & 17.04 & 40.70 & 29.82 & 20.25 & 38.14 & 28.63 & 17.95 \\
Qwen3.5-397B-A17B    & 41.60 & 0.84  & 21.74 & 38.95 & 38.95 & 22.17 & 40.84 & 11.71 & 21.86 \\
\midrule
\multicolumn{10}{c}{\textit{Specialized CUA Model}} \\
\midrule
UI-Tars-1.5-7B      & 10.92 & 0 & 37.38 & 15.79 & 15.79 & 36.43 & 12.31 & 4.5 & 37.11 \\
EvoCUA-8B           & 34.45 & 3.78 & 26.88 & 38.95 & 38.95 & 26.49 & 35.74 & 13.81 & 26.77 \\
EvoCUA-32B          & 37.82 & 13.03 & 27.90 & 47.37 & 46.32 & 21.82 & 40.54 & 22.52 & 26.16 \\
GUI-Owl-1.5-8B      & 44.54 & 37.82 & 20.70 & 42.11 & 31.58 & 22.41 & 43.84 & 36.04 & 21.19 \\
GUI-Owl-1.5-32B     & 47.48 & 38.66 & 24.09 & 49.47 & 47.37 & 24.45 & 48.05 & 41.14 & 24.19 \\
\midrule
\multicolumn{10}{c}{\textit{Ours (based on Qwen3-VL-8B-Instruct)}} \\
\midrule
Qwen3-VL-8B-Instruct & 27.73 & 0.00 & 20.37 & 29.47 & 29.47 & 16.77 & 28.23 & 8.41 & 19.34 \\
\rowcolor{my_purple2}{\textbf{ToolCUA-8B}} & \textbf{45.80} & \textbf{15.13} & \textbf{15.11} & \textbf{49.47} & \textbf{47.37} & \textbf{14.48} & \textbf{46.85} & \textbf{24.32} & \textbf{14.93} \\
\rowcolor{dropred}{ \textbf{$\Delta$} } & +18.07 & +15.13 & -5.26 & +20.00 & +17.90 & -2.29 & +18.62 & +15.91 & -4.41 \\
\bottomrule
\end{tabular}
}
\vspace{-2mm}
\label{table:main-results}
\end{table}

\subsection{Main Results}

\noindent\textbf{Outstanding performance on GUI-Tool Execution Path Selection.}
Table~\ref{table:main-results} summarizes the evaluation results on the OSWorld-MCP benchmark, where ToolCUA-8B achieves state-of-the-art performance of 46.85\% among 8B-class models.
Our model surpasses the previous state-of-the-art GUI-Owl-1.5-8B (43.84\%) and outperforms prominent general foundation models, including Gemini-3.1-Pro (41.14\%) and Claude-4-Sonnet (43.54\%), while trailing the top-tier Claude-4.5-Sonnet by 1.50 percentage points.
This approximately 66\% relative improvement over the baseline (28.23\%) underscores the efficacy of our synthesis-driven scaling pipeline and staged training paradigm.
Beyond task accuracy, ToolCUA also substantially improves GUI-Tool orchestration and execution efficiency. As shown in Table~\ref{tab:mcptool_coordination_failure}, ToolCUA-8B gains 3.9 percentage points over the pure-GUI setting.
Also, compared with the Qwen3-VL-8B-Instruct baseline, ToolCUA increases the overall TIR from 8.41\% to 24.32\%, while reducing ACS from 19.34 to 14.93.
Notably, ToolCUA achieves the lowest ACS (14.93 steps) among all evaluated models, indicating that it not only completes more tasks but also finds more efficient GUI-Tool execution paths with our Tool-Efficient Path Reward.

\begin{figure*}[ht]
  \centering
  \includegraphics[width=0.98\textwidth]{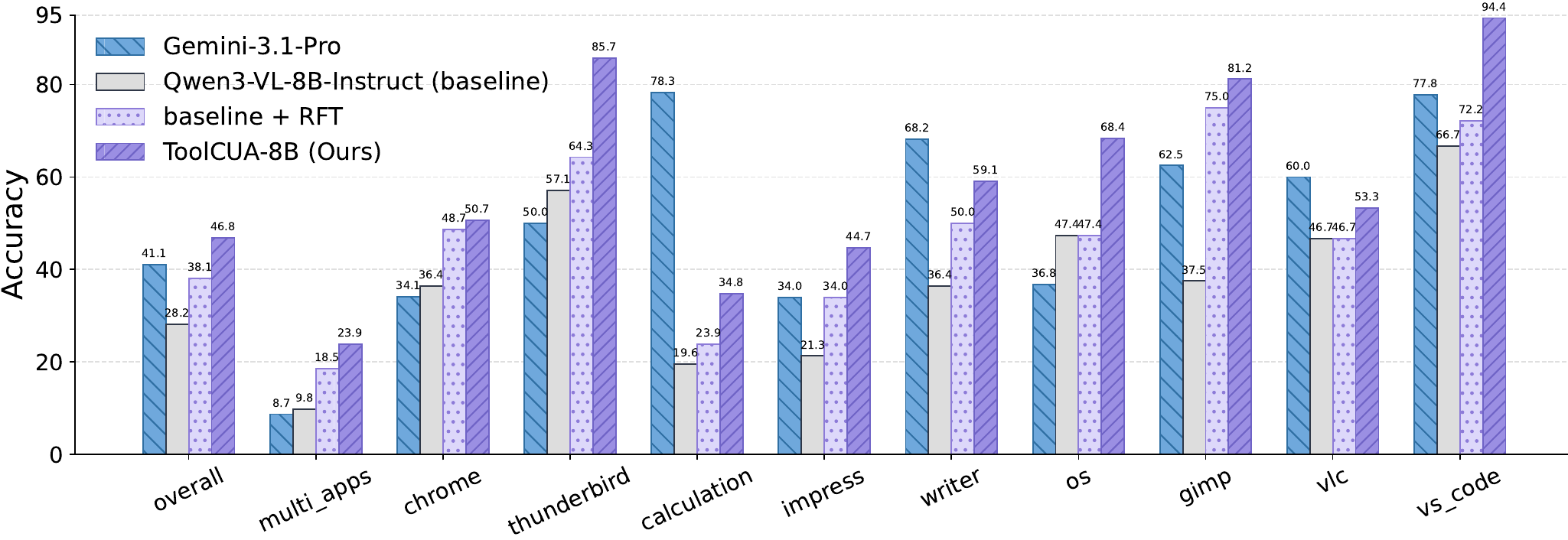}
  \vspace{-2mm}
  \caption{Results across OSWorld-MCP application domains for Gemini-3.1-Pro, Qwen3-VL-8B-Instruct (baseline), the RFT checkpoint ($\mathcal{M}_{rft}$), and ToolCUA-8B.}
  \label{fig:apps_results}
  \vspace{-2mm}
\end{figure*}

\noindent\textbf{Cross-task and Cross-platform Generalization.} ToolCUA also demonstrates strong generalization beyond the training distribution. Although online agentic RL is conducted only on single-application Linux tasks and excludes the \textit{multi\_apps} category, ToolCUA reaches 23.9\% on this held-out domain, improving over both the pre-online-RL checkpoint (18.5\%) and the baseline (9.8\%), as shown in Figure~\ref{fig:apps_results}.
%
%
It also achieves consistent gains across specialized domains, increasing performance from 19.6\% to 34.8\% on \textit{libreoffice\_calculation} and from 66.7\% to 94.4\% on
\textit{vs\_code}.
Beyond cross-task transfer, ToolCUA further generalizes to unseen Windows desktop environments, reaching 33.8\% on the full GUI-only WAA evaluation. For a controlled hybrid-action comparison, Table~\ref{tab:eval_waa} evaluates both GUI-only WAA and WAA with mcp tool support on the same 107-task subset. ToolCUA reaches 37.7\% with GUI actions and improves further to 40.5\% with GUI-Tool actions, whereas the Qwen3-VL-8B-Instruct baseline decreases from 28.0\% to 26.8\% after tools are introduced. This shows that ToolCUA transfers not only its GUI capability but also its learned GUI-Tool orchestration to a new platform and tool ecosystem.

\begin{figure*}[ht]
  \centering
  \includegraphics[width=1.0\textwidth]{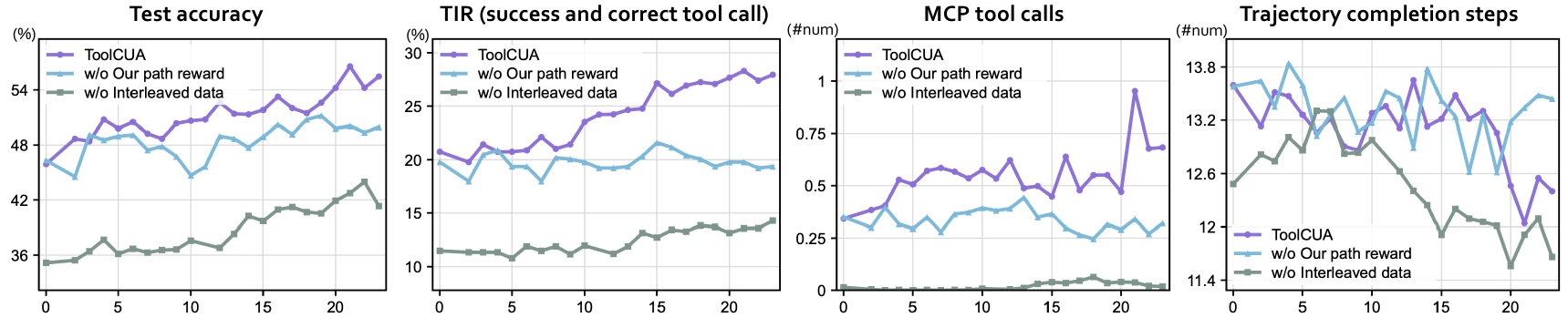}
  \vspace{-2mm}
  \captionof{figure}{Online Agentic RL training dynamics of ToolCUA and two ablations.
  \textit{w/o Interleaved data} removes the offline interleaved GUI-Tool data bootstrapping and directly performs online agentic RL from the baseline model. \textit{w/o Our path reward} replaces the tool-efficient path reward with vanilla GRPO during online agentic RL.}
  \label{fig:rl_dynamics}
  \vspace{-2mm}
\end{figure*}

\subsection{Ablation Analysis}

\noindent
\textbf{The Importance of the Interleaved GUI-Tool Trajectory Data.}
As shown by the ``w/o Interleaved data'' setting in Figure~\ref{fig:rl_dynamics}, we remove cold-start RFT training with our synthetic interleaved GUI-Tool data and directly apply online agentic RL with the tool-efficient path reward to the Qwen3-VL-8B-Instruct baseline model.
Although the model still improves its overall task accuracy during agentic RL, it struggles to acquire reliable tool-calling behavior without the offline interleaved supervision produced by our trajectory scaling pipeline.
Specifically, its TIR remains consistently low and only reaches around 15\% by the end of training, while the number of tool calls stays close to zero throughout most of the learning process.
This suggests that online RL with tool-efficiency rewards alone is insufficient to overcome the GUI-centric bias of base models.
Instead, our data scaling pipeline synthesizes simulated tools into static GUI data to provide grounded, interleaved supervision. This instills diverse tool knowledge and switching priors, establishing a superior foundation for subsequent online exploration.


%

\noindent\textbf{Advantages of Tool-Efficient Path Reward.}
As illustrated by the ``w/o Our path reward'' setting in Figure~\ref{fig:rl_dynamics}, we further train the RFT-initialized model with vanilla multi-turn GRPO, using only the standard $R_{acc}$ and $R_{fmt}$ rewards.
Without the tool-appropriate and trajectory-length reward, the accuracy curve becomes less stable than ToolCUA, showing a clear drop around steps 8--11 and an eventual gap of about 7 percentage points after 20 training steps.
Moreover, both TIR and tool-calls fluctuate without a consistent upward trend, suggesting that the model does not learn when tool invocation is genuinely beneficial.
The trajectory length also lacks a stable downward trend and rebounds after temporary decreases, indicating that vanilla agentic RL does not reliably discover more efficient execution paths.
These results validate that the Tool-Efficient Path Reward is essential for turning online exploration into tool-appropriate and efficiency-aware GUI-Tool path selection.

\begin{table}[!t]
    \centering
    \begin{minipage}[t]{0.47\linewidth}
        \centering
        \small
        \caption{Comparison of pure GUI training and training with hybrid GUI-Tool action spaces.}
        \label{tab:action_space_comparison}
        \resizebox{0.95\linewidth}{!}{%
            \begin{tabular}{llc}
                \toprule
                \textbf{Action} & \textbf{Agent Model}  & \textbf{Acc (\%)} \\
                \midrule
                \multirow{3}{*}{GUI} & Qwen3-VL-8B-Instruct  & 29.03 \\
                & + SFT  & 34.93 \\
                & + SFT + Agentic RL & 42.05 \\
                \midrule
                \multirow{2}{*}{GUI+Tool} &  + RFT  &38.13 \\
                & \textbf{ToolCUA-8B}  & \textbf{46.85} \\
                \bottomrule
            \end{tabular}%
        }
    \end{minipage}%
    \hfill
    \begin{minipage}[t]{0.52\linewidth}
        \centering
        \small
        \caption{Cross-platform evaluation on the same 107-task subset of WindowsAgentArena~\cite{bonatti2024windows} under the GUI-only setting and the setting with mcp tool support. Results are reported as average@3. Tool calls are averaged over successful tasks in the latter setting.}
        \label{tab:eval_waa}
        \resizebox{0.98\linewidth}{!}{%
            \begin{tabular}{lccc}
                \toprule
                \textbf{Agent Model} & \textbf{WAA} & \textbf{WAA + mcp tools} & \textbf{Tool Calls} \\
                \midrule
                Qwen3-VL-8B-Inst. & 28.0 & 26.8 & 0.003 \\
                Qwen3-VL-32B-Inst. & 34.4 & 33.1 & 0.05 \\
                \textbf{ToolCUA-8B} & \textbf{37.7} & \textbf{40.5} & 1.74 \\
                \bottomrule
            \end{tabular}%
        }
    \end{minipage}
    \vspace{-2mm}
\end{table}

\noindent\textbf{Hybrid GUI-Tool Training Is More Effective than Pure-GUI Training.}
We compare hybrid GUI-Tool training with a pure GUI training pipeline in Table~\ref{tab:action_space_comparison}.
For the pure GUI setting, we first cold-start the baseline using a GUI-only dataset of comparable scale to ToolCUA's warmup data, and then conduct agentic RL entirely within the pure GUI action space.
Although GUI-only training improves the baseline from 29.03\% to 34.93\% after SFT and further to 42.05\% after agentic RL, both stages remain below their GUI-Tool counterparts.
In comparison, RFT training with our synthetic interleaved GUI-Tool trajectories already reaches 38.13\%, and the full ToolCUA further improves to 46.85\% after online agentic RL.
These results indicate that the hybrid GUI-Tool action space provides a more effective training environment than pure GUI actions, enabling the agent to learn not only visual grounding but also when structured tool calls can replace redundant low-level operations.

\subsection{Case Study}

As illustrated in Table~\ref{tab:appendix_case1_calculation}, the agent is tasked with creating two pivot tables in a new sheet named Sheet2 to summarize revenue by product and sales channel. While standard GUI-only approaches involve a laborious and error-prone sequence of data range selection and menu navigation, ToolCUA leverages high-level tools for superior efficiency. Specifically, it first retrieves sheet information such as column values and data fields via \textit{libreoffice\_calc.env\_info(Sheet1)} and then identifies the corresponding data range needed for each pivot table. After that, the agent invokes \textit{libreoffice\_calc.create\_pivot\_table(source\_sheet, table\_name, ...)} with the corresponding parameters to generate the tables directly, bypassing brittle step-by-step GUI interactions.
Furthermore, ToolCUA demonstrates the ability to navigate optimal execution paths within hybrid GUI-Tool action spaces. As shown in Table~\ref{tab:appendix_case2_vscode}, the agent successively invokes the efficient \textit{osworld\_mcp\_code.add\_folder} tool twice to add the required directories to the workspace. It then correctly identifies the need for a GUI action and clicks the ``I trust the authors'' dialog to grant folder permissions, thereby successfully completing the end-to-end workflow.
See the complete case study in Appendix~\ref{sec:appendix_case_study}.

\section{Related Work}
\label{sec:related_work}

\noindent\textbf{Multimodal Agents for Computer Use.} Building generalist multimodal agents in digital environments has long been a foundational yet challenging pursuit~\cite{bai2025qwen3, openai2025operator, anthropic2025claudeopus45, xie2024osworld, wang2025opencua, wang2025mobile, ye2025mobilev3, qin2025ui}. 
Current CUAs generally follow two paradigms: multi-agent systems~\cite{agashe2025agent, song2025coact, gonzalez2025unreasonable_agents3, liu2025pc,yang2025gta1,yang2026symphony}, which decompose tasks across specialized modules, and end-to-end agentic models~\cite{qin2025ui, wang2025opencua, liu2025scalecua, xue2026evocua, xu2026mobilev35}, which integrate planning and grounding within a unified policy.
Most of these agents, however, still rely primarily on primitive GUI actions such as clicking, typing, and scrolling, making long-horizon tasks vulnerable to cascading errors and inefficient execution.
Reinforcement learning has therefore been introduced to improve GUI agents, from mobile navigation settings~\cite{lu2025uir1, luo2025guir1} to broader agentic GUI environments~\cite{xu2025mobilerl, yang2025zerogui, lu2025arpo, li2025dartgui, wang2026rlanything}.
These studies demonstrate the promise of environment-driven optimization, but their action spaces are largely restricted to GUI operations and often provide limited supervision for trajectory-level orchestration.
Consequently, they do not directly address the hybrid action spaces that combine raw GUI interaction with structured tool invocation in realistic computer-use environments~\cite{ye2026claw, openclaw2025}.
ToolCUA builds on this direction by using a staged training paradigm to move beyond GUI-only control and optimize complete GUI-Tool execution paths.

\noindent\textbf{Hybrid GUI-Tool Actions for CUAs.} A broad line of research equips LLMs with tools~\cite{qin2023toolllm, patil2023gorilla}, enabling progress in autonomous coding~\cite{dong2025tool, feng2025retool, wang2025acting}, deep research~\cite{team2025tongyi, ye2025agentfold}, and multimodal visual search~\cite{su2025thinking, zheng2025deepeyes, hong2025deepeyesv2, liu2025visual, wang2025adatooler}.
In CUA scenarios, structured tools can reduce repetitive low-level GUI operations and improve efficiency when reliable interfaces are available.
Recent studies~\cite{yan2025mcpworld,jia2025osworldmcp} have begun to explore this direction by introducing various mcp tools for hybrid-action evaluation, while several GUI-Tool systems~\cite{yan2025step,wang2025ui,yang2025ultracua, song2025coact, gonzalez2025unreasonable_agents3, yang2026symphony} connect GUI agents with APIs, SDKs, external executors, or multi-agent tool routers.
Nevertheless, these efforts still fail to solve two critical challenges.
First, high-quality interleaved GUI-Tool trajectories remain scarce because existing pipelines often rely on costly tool construction, environment engineering, or closed-source data collection.
Second, existing methods provide limited guidance for trajectory-level selection. 
ToolCUA addresses these limitations by synthesizing interleaved GUI-Tool trajectories and adopting a staged training paradigm.


\section{Conclusion}

\label{sec:conclusion}

In this work, we presented \textbf{ToolCUA}, an end-to-end computer-use agent that learns to orchestrate GUI actions and tool calls through optimal GUI-Tool path selection.
We show that simply exposing agents to both GUI actions and tool calls is insufficient, as current models often overuse tools or remain overly GUI-centric, leading to inefficient and brittle execution trajectories.
To address this challenge, ToolCUA first scales interleaved GUI-Tool trajectories from existing GUI data without manually constructing tools, then applies Tool-Bootstrapped GUI RFT to acquire tool-calling knowledge and calibrate critical switching decisions.
Finally, Online Agentic RL with a Tool-Efficient Path Reward optimizes ToolCUA in a GUI-Tool environment, encouraging appropriate tool use and shorter execution paths.
Experiments on OSWorld-MCP show that ToolCUA achieves 46.85\% accuracy, a relative improvement of approximately 66\% over the baseline, and consistently improves over pure GUI action settings.
Together with the cross-task, cross-platform, and hybrid Windows transfer results and the ablation study, these findings suggest that training with a hybrid GUI-Tool action space provides a high-fidelity training paradigm for robust real-world digital agents.

\clearpage

\bibliography{ref}

\clearpage
\section*{Appendix}
\appendix
\renewcommand{\appendixname}{\appendixname~\Alph{section}}

\section{Limitations and Future Work}
\label{sec:appendix_limitation_future}

Although ToolCUA demonstrates the effectiveness of learning from synthesized interleaved GUI-Tool trajectories, our synthesis-driven pipeline is still constrained by the state frames and domain coverage of the original GUI-only trajectories. As a result, the diversity and quality of the synthesized hybrid trajectories are coupled with the breadth, fidelity, and task distribution of the source demonstrations. Our tool-scaling process also depends on the capability of the general model used for synthesis; in our internal trials, replacing stronger proprietary models with Qwen3.5-Plus led to noticeably lower generation efficiency and trajectory quality.
Moreover, the synthesized tools are not tied to a specific concrete implementation, which makes them scalable and potentially more generalizable, but also means that real-world execution still depends on the maturity of available tools and the way tool feedback is organized for the CUA.
Finally, due to the scarcity of open-source GUI-Tool coordination benchmarks for computer-use agents, our main performance evaluation is primarily conducted on OSWorld-MCP, leaving broader benchmark coverage as an important limitation.

Future work should further explore hybrid GUI-Tool action spaces across broader platforms, including desktop, mobile, and web environments, where the balance between atomic GUI operations and high-level tools may vary substantially across interface structures and task types.
Another promising direction is to reduce the dependence of agentic RL on heavy sandbox infrastructure by building lighter, more diverse, and more robust environments with hybrid GUI-Tool actions.
We are also interested in asynchronous RL frameworks that decouple training and inference-time rollout, which may improve the scalability and stability of long-horizon policy optimization for computer-use agents.

\section{Broader Impact and Ethics Statement}
\label{sec:appendix_broader_impacts_ethics}

ToolCUA aims to improve real-world digital automation by enabling computer-use agents to coordinate GUI actions and tool calls more efficiently, which may benefit productivity, accessibility, and repetitive workflow assistance. At the same time, more capable CUAs introduce potential risks, including unauthorized operation, accidental modification of user data, privacy leakage from desktop observations, and misuse for automating harmful or deceptive online activities. Our work focuses on benchmarked sandbox environments and does not grant the agent uncontrolled access to personal accounts, sensitive files, or external services. For real-world deployment, we believe such systems should require explicit user consent, transparent action logging, permission boundaries, and human confirmation for high-impact operations.

\section{Implementation Details}
\label{sec:appendix_imple_details}

\subsection{Preliminary Study: Optimal Path Confusion in Hybrid Action Spaces}
\label{sec:appendix_preliminary_study}

To examine whether current CUAs can identify the optimal GUI-Tool path under tool-conditioned contexts, we conduct a diagnostic comparison between pure GUI execution and hybrid GUI-Tool execution across multiple agents, as reported in Table~\ref{tab:mcptool_coordination_failure}.
All models in the hybrid setting are given access to the same tool interface and tool documentation through the system prompt.
Accuracy is reported over the full feasible task set.
ACS reports the average number of trajectory steps over the full feasible task set, reflecting execution efficiency. In Table~\ref{tab:mcptool_coordination_failure}, Tool-Calls reports the average number of tool calls among successful trajectories in the Tool-Beneficial subset; this conditional statistic directly measures tool-use intensity when tools are useful and the task is completed successfully.

\noindent\textbf{Experimental Settings.} For the results reported in Table~\ref{tab:mcptool_coordination_failure}, we follow two evaluation protocols corresponding to the two action-space settings. For the GUI-only setting, we directly use the official OSWorld-released verified trajectories~\cite{xie2024osworld} for EvoCUA-32B and the Claude baselines, set the maximum horizon to 50 steps, and retain only the feasible tasks. For Qwen3VL-8B-Instruct and Qwen3VL-235B-A22B-Thinking, we conduct the evaluation ourselves, aligned with the official implementations.
For clarity, Qwen3-VL-8B-Instruct is abbreviated as Qwen3VL-8B, and Qwen3VL-235B-A22B-Thinking is abbreviated as Qwen3VL-235B throughout Table~\ref{tab:mcptool_coordination_failure}.
%
For the ``GUI+Tool'' setting, we evaluate the same models directly on OSWorld-MCP, ensuring consistency with the official prompts and message construction protocol. We also use the same maximum horizon of 50 steps and report results only on the feasible task subset.
See more metric details in Appendix~\ref{sec:appendix_imple_details_evaluation}.

The results in Table~\ref{tab:mcptool_coordination_failure} reveal a counter-intuitive finding: simply giving a strong model access to both GUI actions and tools does not reliably improve performance. Instead, the hybrid action space often \emph{confuses} the agent, causing it to deviate from the most effective execution path.
More concretely, Table~\ref{tab:mcptool_coordination_failure} exposes two representative failure modes that align with the intuition illustrated in Figure~\ref{fig:cartoon_fig}: once both action spaces are available, the agent stands at a ``forked road'' and often fails to choose the right branch.

\noindent\textbf{Failure Mode I: tool underuse.}
Some models remain overly GUI-centric even when a short tool call is available.
Qwen3VL-8B is the clearest example: even among successful Tool-Beneficial trajectories, it invokes only 0.01 tools on average, indicating that it almost never leaves the GUI branch.
Yet this conservative behavior does not preserve performance: its accuracy drops from 29.0\% to 28.2\%, while ACS also slightly increases from 19.2 to 19.3.
This suggests that the model is unable to recognize the decision boundary where switching to a structured tool would shorten the path and reduce accumulated GUI errors.

\noindent\textbf{Failure Mode II: tool overuse.}
At the other extreme, stronger or larger models may invoke tools aggressively, but frequent tool usage alone does not translate into better task completion.
For example, Claude-4-Sonnet invokes 4.30 tools on average among successful Tool-Beneficial trajectories, yet its overall accuracy drops from 47.7\% to 43.5\% after tools are added. Qwen3VL-235B, EvoCUA-32B, and Claude-4.5-Sonnet exhibit the same failure from the outcome side: access to tools shortens some trajectories but reduces their overall accuracy by 3.0, 12.1, and 13.5 percentage points, respectively.
These cases indicate that the agent is not learning ``use tools more,'' but rather needs to learn \emph{when} tool usage is appropriate.
Over-switching can prematurely abandon necessary GUI grounding, call tools before the required context is established, or commit the trajectory to a brittle tool-heavy path that is shorter but less reliable.


Taken together, Table~\ref{tab:mcptool_coordination_failure} demonstrates that the core challenge of hybrid GUI-Tool agents is optimal GUI-Tool path selection rather than raw action-space expansion.
The issue is not whether the model can recognize that a tool exists, but whether it can determine if the current state calls for continued GUI grounding, an immediate tool invocation, or a later switch after additional GUI setup.
In other words, hybrid execution introduces a trajectory-level decision problem: a locally plausible action may still lead to a globally inferior path.
This is exactly the confusion abstracted by Figure~\ref{fig:cartoon_fig}, where the agent must choose between the GUI path and the tool path without a reliable internal policy for deciding which route is actually better.
In contrast, ToolCUA is the only model in Table~\ref{tab:mcptool_coordination_failure} that benefits from the hybrid action space: its accuracy improves from 42.9\% to 46.8\%, while ACS drops from 19.4 to 14.9 with 2.11 tool calls on average among successful Tool-Beneficial trajectories.
This result suggests that effective hybrid execution does not require excessive tool usage, but a selective switching policy that calls tools only at beneficial points along the trajectory.
Therefore, these preliminary results motivate our training paradigm: ToolCUA is designed to internalize optimal GUI-Tool path selection from interleaved supervision and trajectory-level reinforcement, rather than relying on zero-shot prompting to resolve the fork on its own.

\subsection{Detailed Analysis of Adaptive Tool Usage}
\label{sec:appendix_tool_usage}

The Tool-Calls column in Table~\ref{tab:mcptool_coordination_failure} focuses on successful Tool-Beneficial trajectories. To explain this conditional statistic and distinguish selective invocation from indiscriminate tool use, Table~\ref{tab:appendix_tool_usage_breakdown} further separates tool calls by task type and outcome. ToolCUA concentrates its calls on successful Tool-Beneficial trajectories, averaging 2.11 calls, while using only 0.12 calls on failed Tool-Beneficial trajectories and almost none on Non-Tool-Beneficial tasks. By contrast, Claude-4-Sonnet continues to invoke tools frequently in failed and Non-Tool-Beneficial trajectories. This pattern shows that ToolCUA's tool use is associated with successful, appropriate execution rather than uniformly increased calling frequency.

\begin{table}[ht]
\centering
\small
\caption{Average tool calls separated by task type and outcome on OSWorld-MCP.}
\vspace{1.5mm}
\label{tab:appendix_tool_usage_breakdown}
\resizebox{0.9\linewidth}{!}{%
\begin{tabular}{lccccc}
\toprule
\multirow{2}{*}{\textbf{Model}} & \multicolumn{2}{c}{\textbf{Tool-Beneficial Set (238)}} & \multicolumn{2}{c}{\textbf{Non-Tool-Beneficial Set (95)}} & \multirow{2}{*}{\textbf{ACS $\downarrow$}} \\
\cmidrule(lr){2-3} \cmidrule(lr){4-5}
& {Successful} & {Failed} & {Successful} & {Failed} & \\
\midrule
Kimi-K2.5 & 1.90 & 1.33 & 0.90 & 0.80 & 19.8 \\
Claude-4-Sonnet & 4.30 & 5.40 & 2.90 & 3.80 & 19.8 \\
Qwen3VL-8B & 0.01 & 0.00 & 0.00 & 0.00 & 19.3 \\
\textbf{ToolCUA-8B} & \textbf{2.11} & 0.12 & 0.02 & 0.00 & \textbf{14.9} \\
\bottomrule
\end{tabular}%
}
\end{table}

Averaged over all 333 feasible tasks, ToolCUA makes 0.74 tool calls per task; this pooled value is reduced by the 95 Non-Tool-Beneficial tasks, where abstention is the desired behavior. The conditional breakdown therefore provides a more direct view of orchestration quality. For an empirical efficiency reference, we pool all evaluated models, retain the shortest successful tool-invoking trajectory for each covered task, and average over Tool-Beneficial tasks. This reference covers the union of 249 tasks and averages 2.65 tool calls; it should not be interpreted as a ground-truth optimum. ToolCUA's 2.11 calls on successful Tool-Beneficial trajectories are close to this empirical reference, while its near-zero usage elsewhere indicates that it neither remains GUI-only nor invokes tools indiscriminately.

\subsection{Data Statistics of the Scaling Pipeline}
\label{sec:appendix_imple_details_data}

Our Offline Interleaved GUI-Tool Trajectory Scaling Pipeline constructs a robust foundation by aggregating diverse pure GUI interaction trajectories from OpenCUA~\cite{wang2025opencua} and ScaleCUA~\cite{liu2025scalecua}.
To further augment the data, we leverage diverse powerful MLLMs to execute sandbox rollouts across internal tasks, resulting in 1,200 curated, high-quality trajectories post-filtering.
As detailed in Table~\ref{tab:appendix_data_source}, this combined source corpus comprises 10,000 trajectories and 192,000 raw GUI steps, with OpenCUA serving as the primary source (8,500 trajectories). From this base, our pipeline synthesizes 10,000 interleaved GUI-Tool trajectories, yielding 180k high-quality steps for the warmup Supervised Fine-Tuning (SFT) stage. Furthermore, we extract a specialized subset of 5k critical switching steps ($\mathcal{D}_{critical}$) to facilitate offline single-turn Reinforcement Learning (RL).

The synthesis process generates a diverse and hierarchically structured tool inventory. As outlined in Table~\ref{tab:appendix_synthesis_data}, the synthetic dataset contains 4,350 unique tools spanning multiple abstraction levels. It is primarily anchored by fine-grained and mid-grained operations, complemented by coarse-grained skills. To emulate realistic decision-making complexity, each trajectory provides an average tool pool of 19.75 candidate tools, from which the agent executes an average of 7.89 tools. Overall, these metrics show that our synthesized trajectories cover diverse tool types, large candidate pools, and substantial executed tool calls.

\noindent\textbf{Human Verification of Next-State Grounding.} Overall, 82.5\% of synthesized tool steps pass the automatic next-state grounding filter. We further audit both accepted and rejected examples manually. For each sampled tool step, an annotator examines the tool description, schema, call arguments, and corresponding next-state screenshot, assumes normal execution according to the tool specification, and judges whether the predicted effect is aligned with the recorded GUI state. As shown in Table~\ref{tab:appendix_grounding_audit}, 93.0\% of accepted steps are judged aligned, whereas only 28.0\% of rejected steps are judged aligned. This contrast indicates that the filter retains state-consistent synthetic steps while removing most low-consistency candidates.

\begin{table}[ht]
\centering
\small
\caption{Human audit of the next-state grounding filter. State alignment denotes the percentage of sampled tool steps whose predicted effects are judged consistent with the recorded next-state screenshot.}
\label{tab:appendix_grounding_audit}
\vspace{1.5mm}
\begin{tabular}{lcc}
\toprule
\textbf{Tool-step set} & \textbf{Sampled} & \textbf{State alignment $\uparrow$} \\
\midrule
Accepted by next-state grounding & 100 & 93.0\% \\
Rejected by next-state grounding & 50 & 28.0\% \\
\bottomrule
\end{tabular}
\end{table}

\begin{figure}[ht]
  \centering
  \small
  \includegraphics[width=0.9\textwidth]{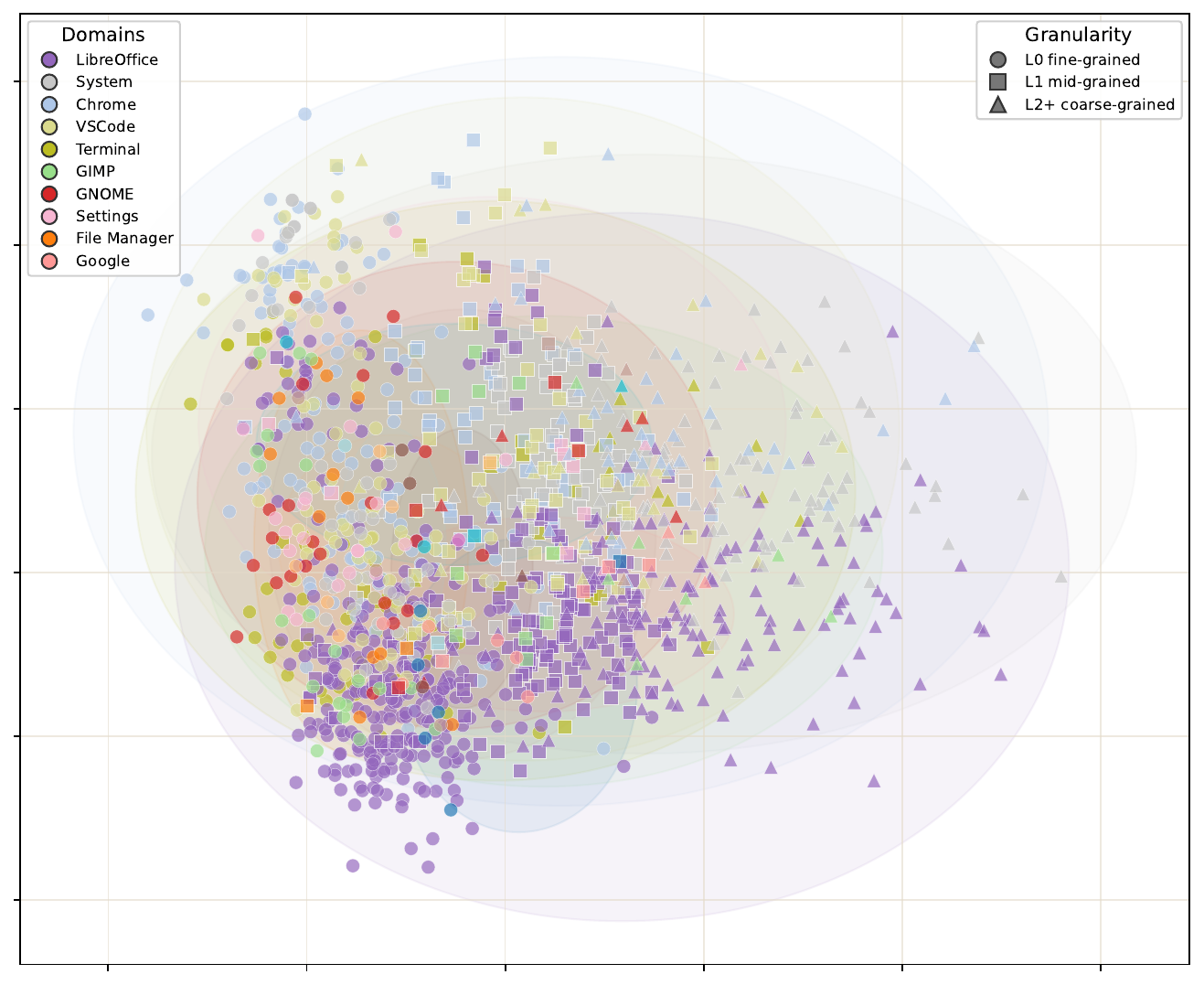}
  \vspace{-2mm}
  \caption{Visualization of the synthesized tools in a projected action space, where each point corresponds to one tool node, colors denote the application taxonomy, and marker shapes denote granularity tiers. }
\label{fig:tool-domain-map}
  \vspace{-2mm}
\end{figure}

Figure~\ref{fig:tool-domain-map} visualizes the synthesized tools in a projected space, where each point corresponds to one tool node and marker shapes denote granularity tiers. Major application categories such as \texttt{LibreOffice}, \texttt{Chrome}, and \texttt{VSCode} occupy broad and distinguishable regions, indicating that the synthesized inventory does not collapse into a narrow set of routines. Within each region, the coexistence of fine-grained, mid-grained, and coarse-grained tools shows that the learned tool space is not restricted to a single abstraction level. The visualization therefore supports the two intended properties of our bottom-up pipeline: diverse tool functionality and clear multi-granular organization.

\begin{table}[ht]
    \begin{minipage}[t]{0.5\linewidth}
       \caption{Data distribution of pure GUI trajectories used by our synthesis pipeline.}
        \label{tab:appendix_data_source}
      \centering
    \resizebox{0.98\linewidth}{!}{
      \centering
        \begin{tabular}{lrr}
        \toprule
        \textbf{Data Source} & \textbf{Trajs} & \textbf{Steps} \\
        \midrule
        OpenCUA~\cite{wang2025opencua} & 8500 & 110k \\
        ScaleCUA~\cite{liu2025scalecua}& 300 & 20k \\
        Sandbox collected & 1200 & 62k \\
        \midrule
        \textbf{Source total} & \textbf{10k} & \textbf{192k} \\
        critical steps ($\mathcal{D}_{critical}$) & - & 5k \\
        \bottomrule
        \end{tabular}
      }
    \end{minipage}
    \hfill
    \begin{minipage}[t]{0.48\linewidth}
      \centering
        \caption{Statistics of our interleaved GUI-Tool data.}
        \label{tab:appendix_synthesis_data}
    \resizebox{0.98\linewidth}{!}{
        \begin{tabular}{lr}
        \toprule
        \small
        \textbf{Statistics} & \textbf{Count} \\
        \midrule
        Unique Tools & 4,350 \\
        fine-grained &  2,000 \\
        mid-grained &  1,900 \\
        coarse-grained &  450 \\
        avg. tool-pool size per traj. & 19.75 \\
        avg. executed tools per traj. & 7.89 \\
        \bottomrule
        \end{tabular}
    }
        \end{minipage}
\end{table}

\subsection{Training Details}
\label{sec:appendix_imple_details_training}

\noindent\textbf{Tool-Bootstrapped GUI RFT.}
In our staged training paradigm, we adopt Qwen3-VL-8B-Instruct~\cite{bai2025qwen3} as our base model $M$.
The warmup SFT is conducted for 3 epochs using full-parameter fine-tuning of both the vision tower and the LLM backbone on a cluster of $8 \times 8$ GPUs, resulting in the intermediate model $\mathcal{M}_{sft}$.
For the offline single-turn RL phase, we use a rollout group size of 32, a learning rate of $1 \times 10^{-6}$, and a training batch size of 128 to produce $\mathcal{M}_{rft}$.

\noindent\textbf{Online Agentic RL Training.} During the subsequent online agentic RL stage, we set hyperparameters $\lambda = 0.4$, $\beta = 0.2$, and $S_{max} = 30$ to define the maximum execution horizon. The training configuration for this stage includes a rollout size of 32 per group, a learning rate of $1 \times 10^{-6}$, and a training batch size of 32 to obtain \textbf{ToolCUA}.
\noindent\textbf{Tool-Beneficial Labeling.} For the Tool Appropriateness Reward Term, we assign a task-level label $t_b \in \{1,-1\}$ through rollout-based comparison followed by human verification. A frontier MLLM performs rollouts for each task with and without mcp tools using pass@32. We label a task as tool-beneficial when the with-tool success rate clearly exceeds the GUI-only success rate, and as non-tool-beneficial in the opposite case. The OSWorld-MCP annotations are used only as a cross-reference rather than as supervision for the reward.
Human annotators then verify whether the available tools genuinely simplify, accelerate, or improve the reliability of each task. In a sampled audit, rollout-based labels agree with human judgments on 91.7\% of 60 tool-beneficial tasks and 80.0\% of 40 non-tool-beneficial tasks. Cases without a clear success-rate margin are treated as ambiguous and excluded from the online RL pool; these account for fewer than 5\% of the manually checked tasks. Since $R_{tool}$ is a soft shaping term with weight $\lambda=0.4$ and is activated only for successful trajectories, residual label noise does not override the task-success signal.
\noindent\textbf{Online RL Task Pool.} The online RL pool contains 339 single-domain tasks and excludes the entire \textit{multi\_apps} domain, which is reserved for held-out evaluation. Of these tasks, 141 are drawn directly from OSWorld~\cite{xie2024osworld}, 124 are rewritten and augmented variants, and 74 are newly scaled tasks constructed following RLAnything~\cite{wang2026rlanything}. The rewritten tasks vary the original phrasing, user intent, and constraints, while the scaled tasks further alter task queries and difficulty and use newly constructed evaluators. We retain source-task identifiers throughout construction and perform task-ID-level deduplication when organizing the training split.

\noindent\textbf{CUA Agentic RL in a GUI-Tool Environment.} To support large-scale online exploration, we implement a decoupled training-inference infrastructure using the verl~\cite{sheng2024hybridflow} framework, where policy optimization occurs on a GPU cluster while rollouts are executed on distributed ECS servers. The environment sandbox is built upon OSWorld~\cite{xie2024osworld} QEMU images, incorporating mcp tool designs from OSWorld-MCP~\cite{jia2025osworldmcp} and AutoGLM~\cite{liu2024autoglm}.
We further optimize the tool-calling interface by designing an agent-readable return format that provides concise, semantically dense feedback to reduce token overhead and improve grounding accuracy. %
The online RL phase uses around 250 independent Docker instances for environment rollouts, together with $8 \times 8$ GPUs for policy training and $4 \times 8$ GPUs for dedicated inference serving.

\subsection{Benchmark Evaluation}
\label{sec:appendix_imple_details_evaluation}

We evaluate ToolCUA primarily on OSWorld-MCP~\cite{jia2025osworldmcp}, which extends OSWorld with tool actions and therefore directly measures hybrid GUI-Tool execution. Following the benchmark protocol, we report three metrics: task accuracy, Tool Invocation Rate (TIR), and Average Completion Steps (ACS). To reduce sandbox stochasticity, we report average@3 results for OSWorld-MCP and set the maximum number of execution steps to 50 for each task.

\textbf{Task Accuracy. }
Task accuracy is the primary success metric and measures whether an agent completes the target instruction according to the benchmark evaluator. In OSWorld-MCP, this metric reflects not only visual grounding and GUI interaction ability, but also whether the agent can use the provided tools to reach the correct final state.

\textbf{Tool Invocation Rate (TIR). }
OSWorld-MCP further separates tasks into Tool-Beneficial Tasks and Non-Tool-Beneficial Tasks, allowing us to measure whether an agent invokes tools in appropriate contexts rather than merely calling tools more often. Let $N_t$ be the total number of Tool-Beneficial Tasks, and $n_t$ the number of such tasks in which the agent invoked a tool and successfully completed the task during evaluation. Let $N_g$ be the total number of Non-Tool-Beneficial Tasks, and $n_g$ the number of such tasks in which the agent did not invoke a tool and successfully completed the task. We define TIR as:

\begin{equation}
\textrm{TIR} = {(n_t + n_g)}/{(N_t + N_g)}
\end{equation}

TIR therefore captures whether the agent aligns its tool usage with task-level tool utility, including both using tools when they are beneficial and avoiding them when GUI actions are more appropriate.

\textbf{Average Completion Steps (ACS). }
ACS measures the average number of environment interaction steps used by an agent across tasks. For $N$ tasks, if the number of execution steps for task $i$ is $S_i$, the Average Completion Steps is computed as:
\begin{equation}
\textrm{ACS} = {\sum_{i=1}^{N}{S_i}}/{N}
\end{equation}

ACS reflects execution efficiency: agents that identify shorter tool-conditioned paths and avoid redundant GUI operations generally require fewer steps to complete the same task.

To assess cross-platform transfer, we evaluate ToolCUA in both the GUI-only WindowsAgentArena (WAA) setting and a setting with mcp tool support. We port 117 mcp tools covering five domains---\texttt{file\_explorer}, \texttt{libreoffice\_calc}, \texttt{libreoffice\_writer}, \texttt{vlc}, and \texttt{vs\_code}---which correspond to 107 evaluation tasks. Applications whose tool support could not be stabilized, including \texttt{microsoft\_paint}, \texttt{msedge}, and \texttt{notepad}, are excluded from this matched subset. In the setting with mcp tool support, the agent is jointly provided with the standard GUI action space and the available ported mcp tool schemas, following the same message construction and tool-calling interface used in our Linux evaluation. For a controlled comparison, results for the GUI-only setting and the setting with mcp tool support in Table~\ref{tab:eval_waa} are recomputed on the same 107 tasks. We set \texttt{max\_steps}=50 and report average@3 accuracy. Tool calls are averaged over successful trajectories in the setting with mcp tool support.

\subsection{Ablation Details}
\label{sec:appendix_imple_details_ablation}

We conduct three ablation studies to analyze the contribution of each training component and the effect of the hybrid GUI-Tool action space. %
The first two studies, shown in Figure~\ref{fig:rl_dynamics}, examine the effectiveness of our staged training design. %
In the first setting, denoted as ``w/o Interleaved data'', we remove the offline interleaved GUI-Tool data bootstrapping stage and directly apply online agentic RL with the proposed Tool-Efficient Path Reward to the Qwen3-VL-8B-Instruct baseline. %
This setting tests whether online exploration alone can acquire tool-calling knowledge and GUI-Tool switching ability from sparse trajectory-level feedback. %
In the second setting, the model is first initialized with Tool-Bootstrapped GUI RFT, and then optimized with a vanilla GRPO-based agentic RL objective that only uses the standard accuracy reward $R_{acc}$ and format reward $R_{fmt}$. %
This variant removes the Tool Appropriateness Reward Term and Path Efficiency Reward Term, isolating the effect of the Tool-Efficient Path Reward after the model has already acquired basic tool-calling knowledge and local switching capability.

The third ablation, reported in Table~\ref{tab:action_space_comparison}, compares training in the hybrid GUI-Tool action space with training in a pure GUI action space. %
For the pure-GUI setting, we fine-tune the Qwen3-VL-8B-Instruct baseline using a GUI-only dataset of comparable scale to the warmup SFT data, and then perform online agentic RL within the pure GUI action space. %
Since tool actions are unavailable in this environment, the reward consists only of $R_{acc} + R_{fmt} + R_{length}$, where the Path Efficiency Reward Term still encourages efficient execution but cannot reward tool-appropriate decisions. %
The final evaluation is also conducted under the pure GUI action space, allowing us to directly compare whether hybrid GUI-Tool training provides a higher-fidelity optimization setting than GUI-only training.

For all online agentic RL ablations, we follow the evaluation protocol in Section~\ref{sec:appendix_imple_details_evaluation} and report average@3 results. %
The maximum rollout horizon during training is set to \texttt{max\_steps}=30. %
Due to the computational cost of sandbox-based online RL under our decoupled training-inference infrastructure, each run is trained for roughly 25 optimization steps. %
With a unified training batch size of 32 and dynamic filtering enabled, each ablation uses approximately 1,200 effective training samples. %
Each run consumes approximately $8 \times 8$ GPUs together with distributed ECS sandbox workers for about six days.

\newpage
\section{More Case Studies}
\label{sec:appendix_case_study}

\setlength{\LTcapwidth}{\textwidth}  
\begin{longtable}{m{3.5cm}m{3.5cm}m{5.8cm}}
\caption{Case Study: LibreOffice Calc (ID: 535364ea-05bd-46ea-9937-9f55c68507e8).}
\label{tab:appendix_case1_calculation} 
\\
\textbf{Agent Prediction} & \textbf{Tool-Calling Response} & \textbf{Screenshot} \\
\hline
\tiny
\begin{minipage}{.25\textwidth}
\begin{lstlisting}
Step 2: Action: Get the workbook information to identify the file path and sheet details.

<tool_call>
{\"name\": \"osworld_mcp_libreoffice_calc.get_workbook_info\", \"arguments\": {}}
</tool_call>
\end{lstlisting}
\end{minipage} 
& 
\tiny
\begin{minipage}{.25\textwidth}
\begin{lstlisting}
<tool_response>
{"type": "text", "text": "{\"success\": true, \"result\": {\"file_path\": \"file:///home/user/SummerSales.xlsx\", \"file_title\": \"SummerSales.xlsx\", \"sheets\": [{\"name\": \"Sheet1\", \"index\": 0, \"visible\": true, \"row_count\": 30, \"column_count\": 7}], \"active_sheet\": \"Sheet1\", \"sheet_count\": 1}, \"error_message\": null}"}
</tool_response>
\end{lstlisting}
\end{minipage} 
& 
\begin{minipage}{.4\textwidth}
\includegraphics[width=6cm, height=3.3cm]{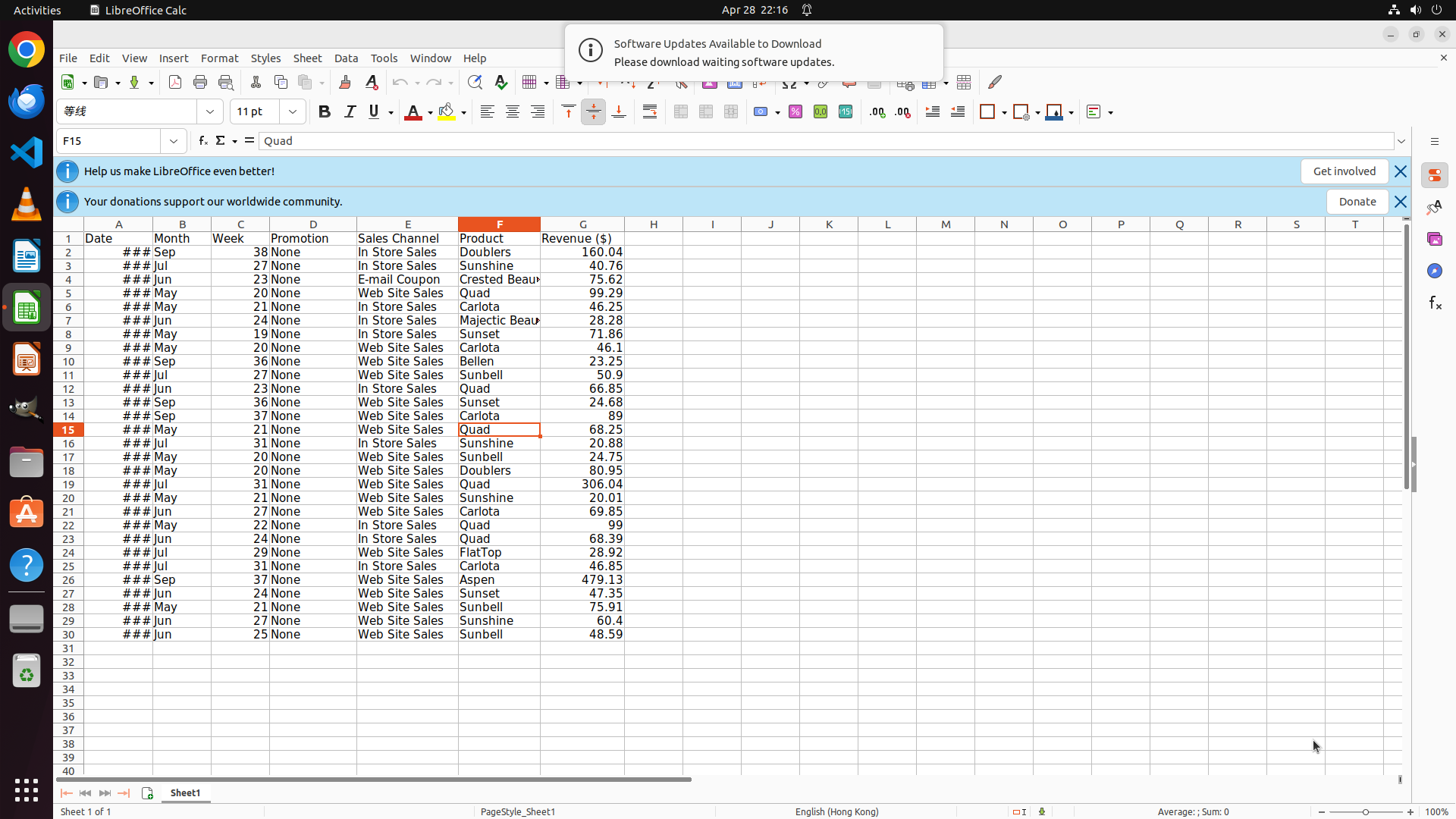}
\end{minipage}\\
\hline
\tiny
\begin{minipage}{.25\textwidth}
\begin{lstlisting}
Step 3: Action: Get content information of Sheet1 to understand the data structure including headers and data ranges.

<tool_call>
{\"name\": \"osworld_mcp_libreoffice_calc.env_info\", \"arguments\": {\"sheet_name\": \"Sheet1\"}}
</tool_call>
\end{lstlisting}
\end{minipage} 
& 
\tiny
\begin{minipage}{.25\textwidth}
\begin{lstlisting}
<tool_response>
{"type": "text", "text": "{\"success\": true, \"result\": {\"name\": \"Sheet1\", \"data\": \"  | A| B| C | D | E | F | G |\\n--|-----...}, \"error_message\": null}"}
</tool_response>
\end{lstlisting}
\end{minipage}  
& 
\begin{minipage}{.4\textwidth}
\includegraphics[width=6cm, height=3.3cm]{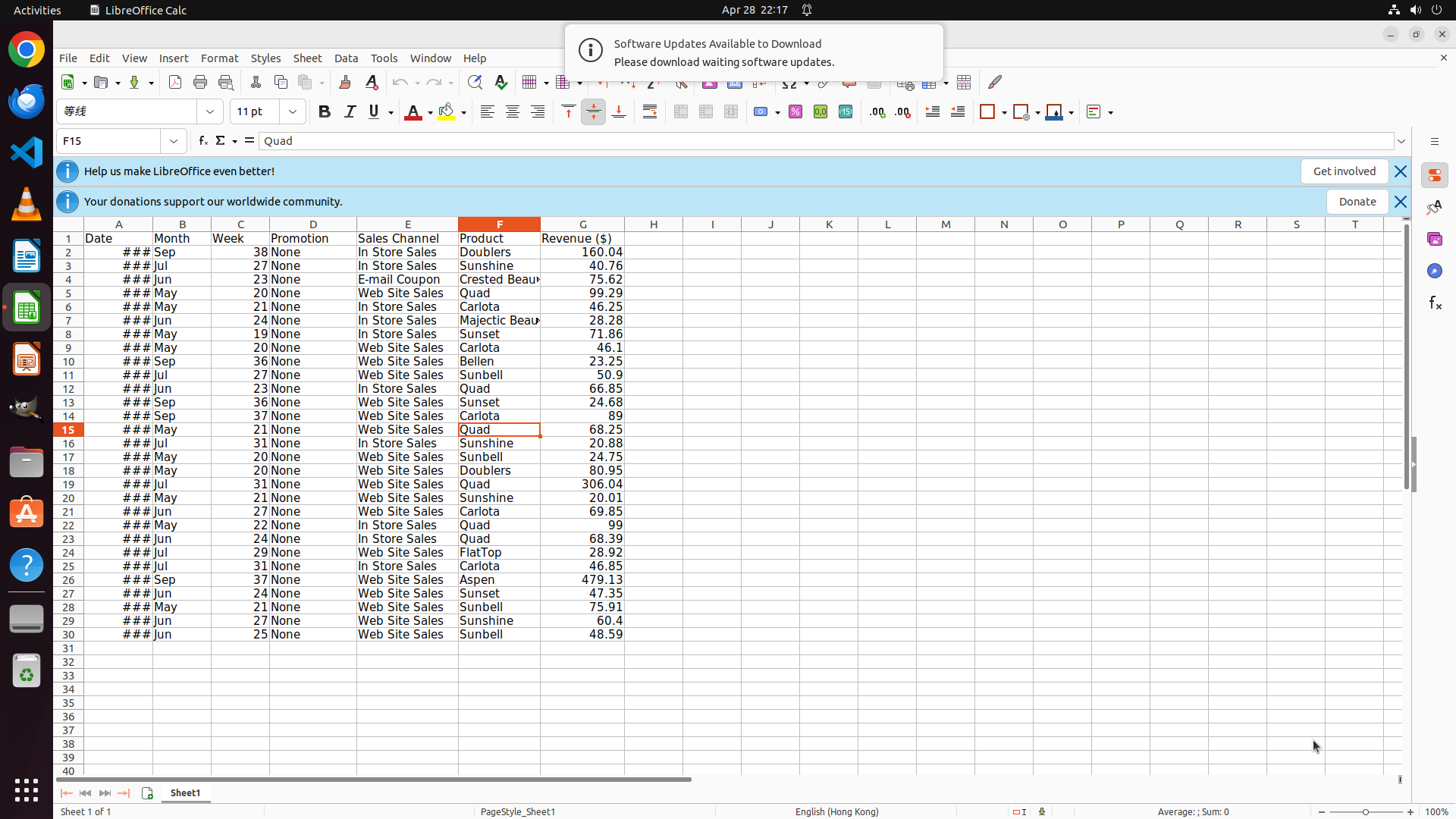}
\end{minipage}\\
\hline
\tiny
\begin{minipage}{.25\textwidth}
\begin{lstlisting}
Step 4: Action: Create a new sheet named \"Sheet2\" to serve as the destination for the pivot tables.

<tool_call>
{\"name\": \"osworld_mcp_libreoffice_calc.switch_active_sheet\", \"arguments\": {\"sheet_name\": \"Sheet2\"}}
</tool_call>
\end{lstlisting}
\end{minipage} 
& 
\tiny
\begin{minipage}{.25\textwidth}
\begin{lstlisting}
<tool_response>
{"type": "text", "text": "{\"success\": true, \"result\": \"Success\", \"error_message\": null}"}
</tool_response>
\end{lstlisting}
\end{minipage} 
& 
\begin{minipage}{.4\textwidth}
\includegraphics[width=6cm, height=3.3cm]{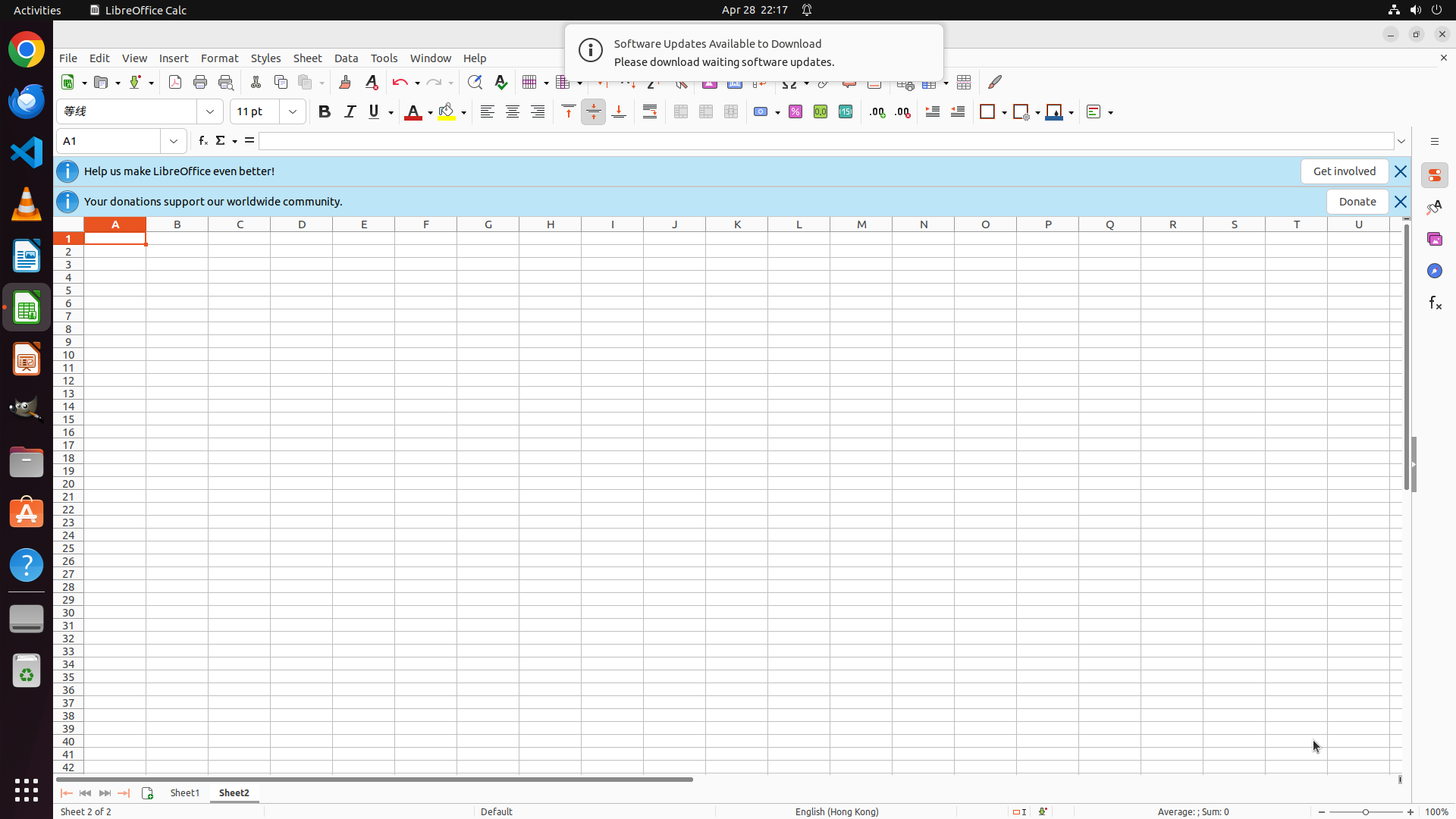}
\end{minipage} \\
\hline
\tiny
\begin{minipage}{.25\textwidth}
\begin{lstlisting}
Step 5: Action: Create the first pivot table named 'PivotTable1' in Sheet2 to show total revenue for each product.
<tool_call>
{"name": "osworld_mcp_libreoffice_calc.create_pivot_table", "arguments": {"source_sheet": "Sheet1", "table_name": "PivotTable1", "row_fields": ["F"], "value_fields": ["G"], "aggregation_function": "sum", "target_cell": "A1"}}
</tool_call>
\end{lstlisting}
\end{minipage} 
& 
\tiny
\begin{minipage}{.25\textwidth}
\begin{lstlisting}
<tool_response>
{"success": true, "result": "Success", "error_message": null}
</tool_response>
\end{lstlisting}
\end{minipage} 
& 
\begin{minipage}{.4\textwidth}
\includegraphics[width=6cm, height=3.3cm]{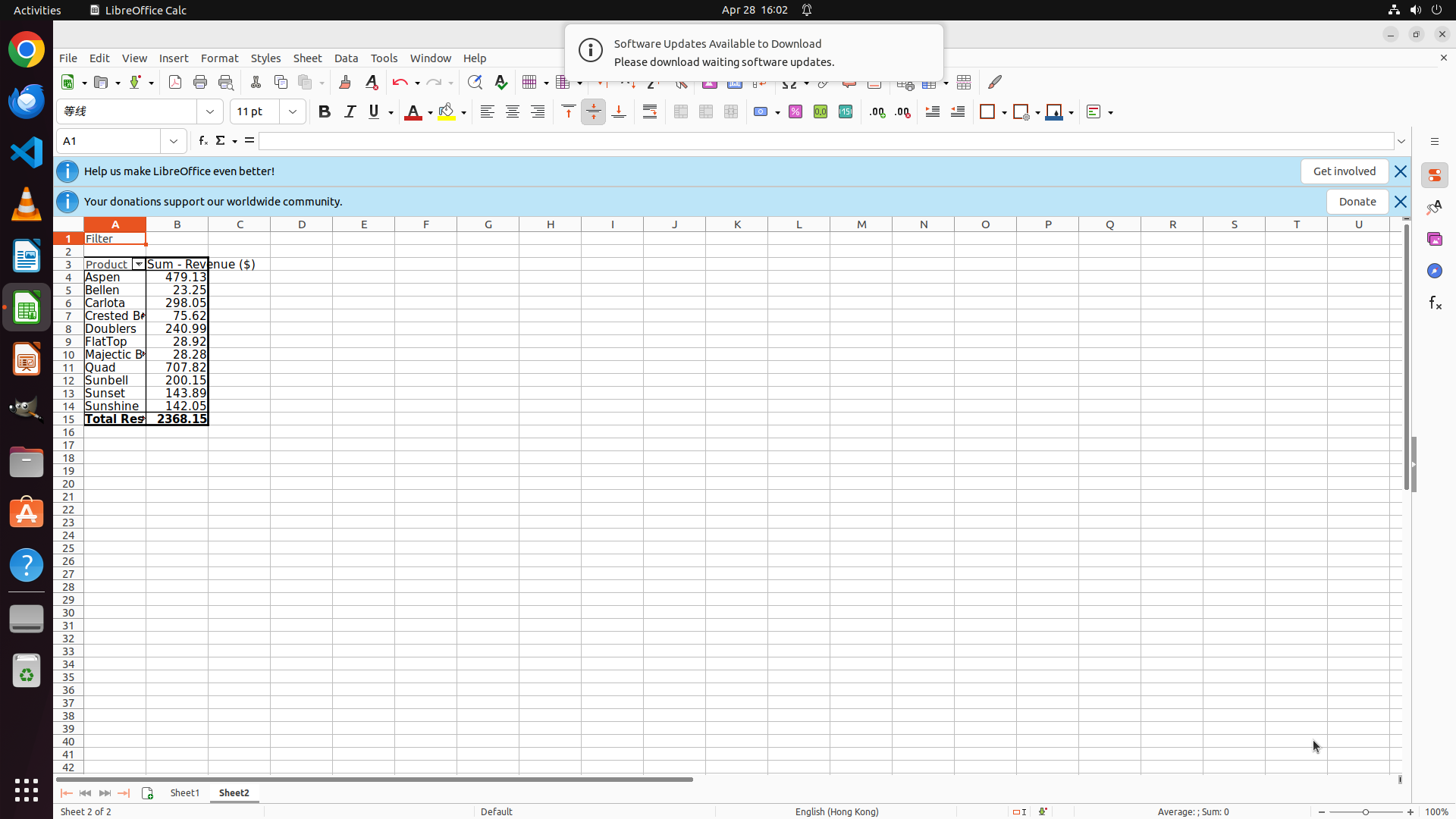}
\end{minipage}\\
\hline
\tiny
\begin{minipage}{.25\textwidth}
\begin{lstlisting}
Step 6: Action: Create the second pivot table named 'PivotTable2' in Sheet2 to show total revenue for each sales channel.
<tool_call>
{"name": "osworld_mcp_libreoffice_calc.create_pivot_table", "arguments": {"source_sheet": "Sheet1", "table_name": "PivotTable2", "row_fields": ["E"], "value_fields": ["G"], "aggregation_function": "sum", "target_cell": "A19"}}
</tool_call>
\end{lstlisting}
\end{minipage} 
& 
\tiny
\begin{minipage}{.25\textwidth}
\begin{lstlisting}
<tool_response>
{"success": true, "result": "Success", "error_message": null}
</tool_response>
\end{lstlisting}
\end{minipage} 
& 
\begin{minipage}{.4\textwidth}
\includegraphics[width=6cm, height=3.3cm]{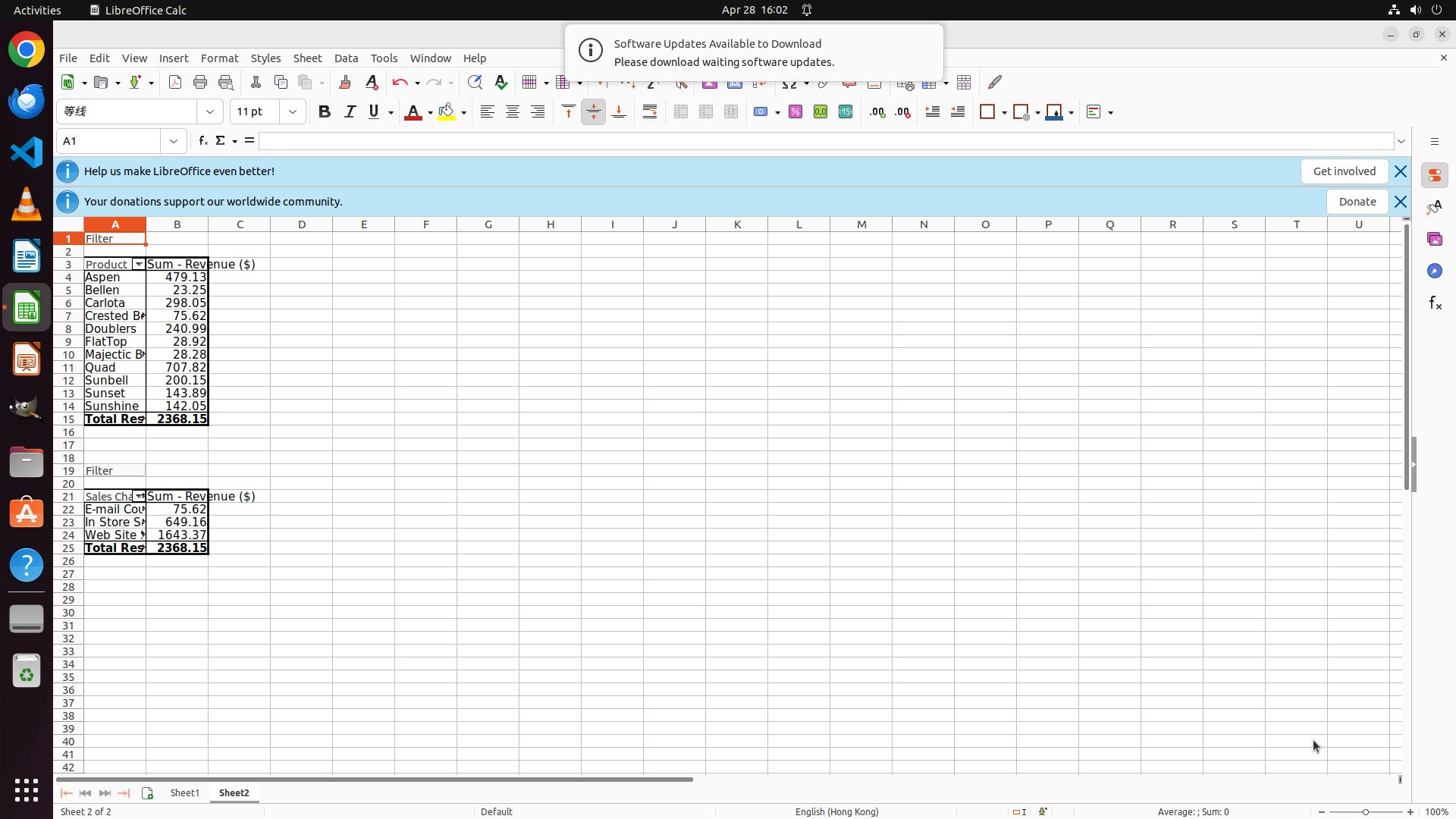}
\end{minipage} \\
\hline
\end{longtable}

\vspace{-4mm}
\textbf{Analysis: } This case examines a LibreOffice Calc task that requires two pivot tables in a new sheet named Sheet2, showing total revenue by product and by sales channel. Tool calls allow the agent to inspect workbook metadata and sheet contents before directly invoking \texttt{create\_pivot\_table} with the fields obtained from the environment information.

\newpage
\setlength{\LTcapwidth}{\textwidth}  
\begin{longtable}{m{3.5cm}m{3.5cm}m{5.8cm}}
\caption{Case Study: VS Code (ID: 6ed0a554-cbee-4b44-84ea-fd6c042f4fe1).}
\label{tab:appendix_case2_vscode} 
\\
\textbf{Agent Prediction} & \textbf{Tool-Calling Response} & \textbf{Screenshot} \\
\hline
\tiny
\begin{minipage}{.25\textwidth}
\begin{lstlisting}
Step 1: Action: Add the folder '/home/user/data1' to the current workspace in VSCode.

<tool_call>
{\"name\": \"osworld_mcp_code.add_folder\", \"arguments\": {\"folder\": \"/home/user/data1\"}}
</tool_call>"
\end{lstlisting}
\end{minipage} 
& 
\tiny
\begin{minipage}{.25\textwidth}
\begin{lstlisting}
<tool_response>
{"type": "text", "text": "{\"success\": true, \"result\": \"Successfully added folder\", \"error_message\": null}"}
</tool_response>
\end{lstlisting}
\end{minipage} 
& 
\begin{minipage}{.4\textwidth}
\includegraphics[width=6cm, height=3.3cm]{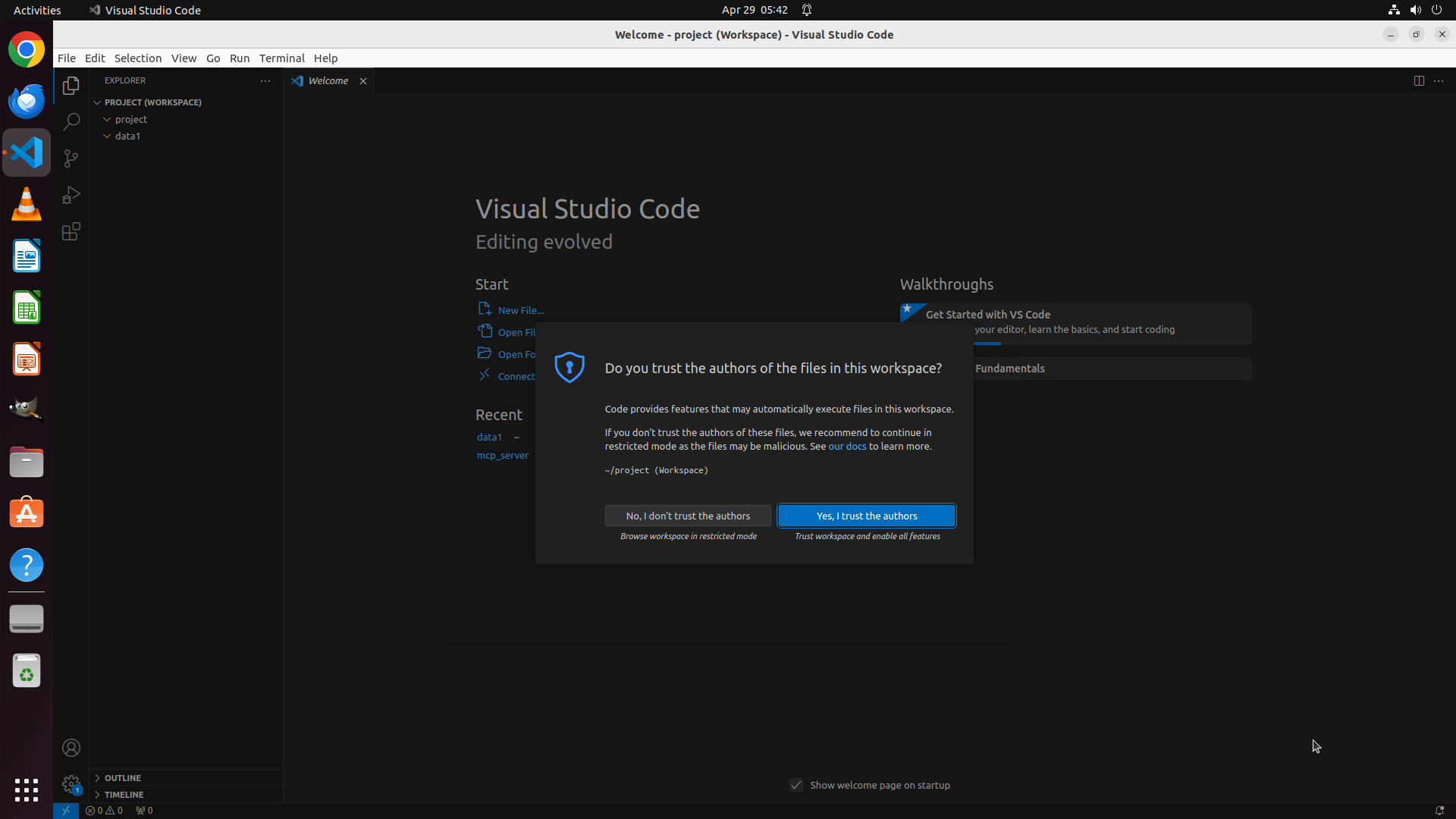}
\end{minipage}\\
\hline
\tiny
\begin{minipage}{.25\textwidth}
\begin{lstlisting}
Step 2: Action: Add the folder '/home/user/data2' to the current workspace in VSCode.

<tool_call>
{\"name\": \"osworld_mcp_code.add_folder\", \"arguments\": {\"folder\": \"/home/user/data2\"}}
</tool_call>"
\end{lstlisting}
\end{minipage} 
& 
\tiny
\begin{minipage}{.25\textwidth}
\begin{lstlisting}
<tool_response>
{"type": "text", "text": "{\"success\": true, \"result\": \"Successfully added folder\", \"error_message\": null}"}
</tool_response>
\end{lstlisting}
\end{minipage}  
& 
\begin{minipage}{.4\textwidth}
\includegraphics[width=6cm, height=3.3cm]{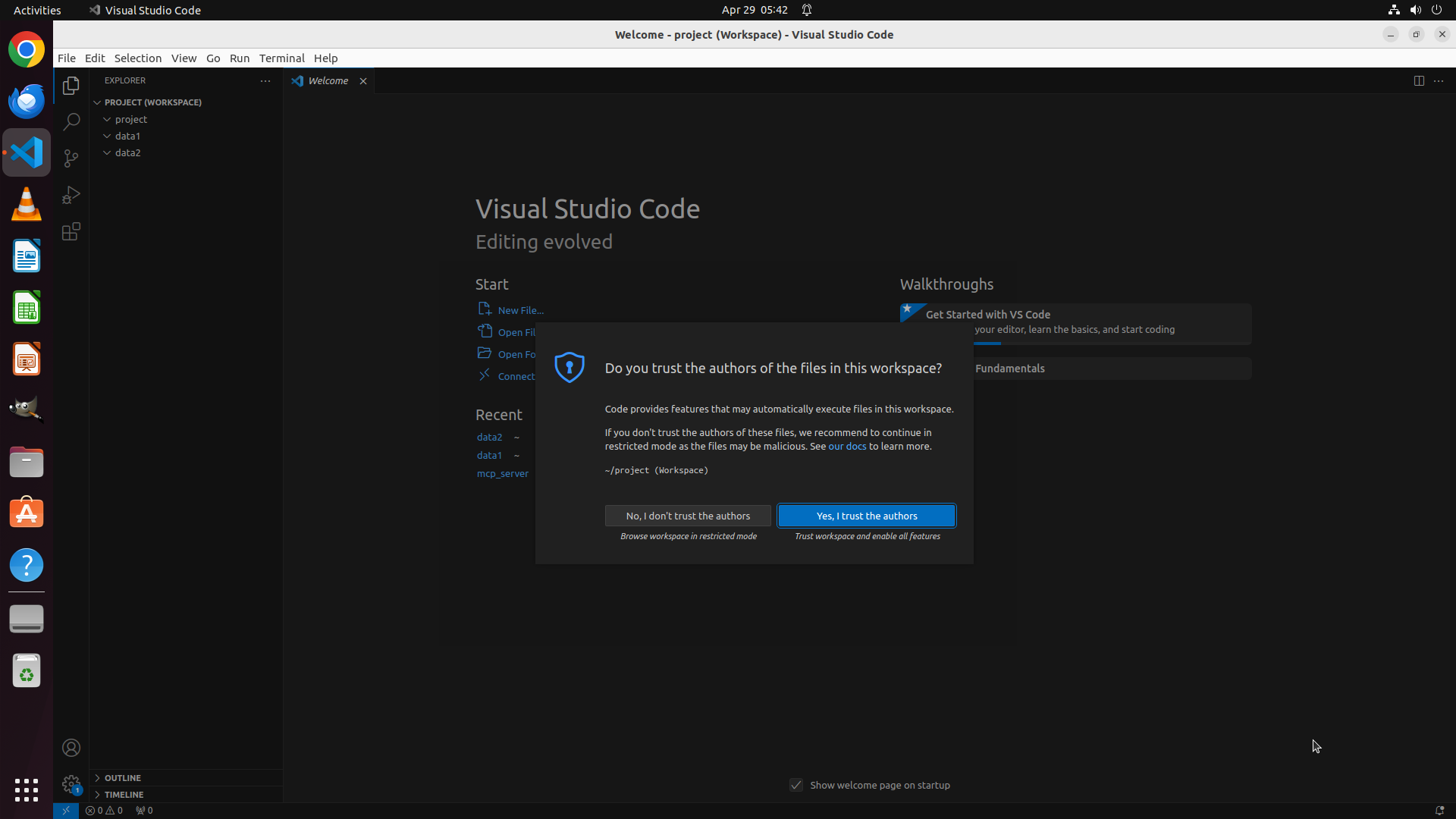}
\end{minipage}\\
\hline
\tiny
\begin{minipage}{.25\textwidth}
\begin{lstlisting}
Step 3: Action: Click on the "Yes, I trust the authors" button in the trust dialog popup.

<tool_call>
{\"name\": \"computer_use\", \"arguments\": {\"action\": \"click\", \"coordinate\": [596, 627]}}
</tool_call>
\end{lstlisting}
\end{minipage} 
& 
\tiny
\begin{minipage}{.25\textwidth}
\begin{lstlisting}
## Not available
\end{lstlisting}
\end{minipage} 
& 
\begin{minipage}{.4\textwidth}
\includegraphics[width=6cm, height=3.3cm]{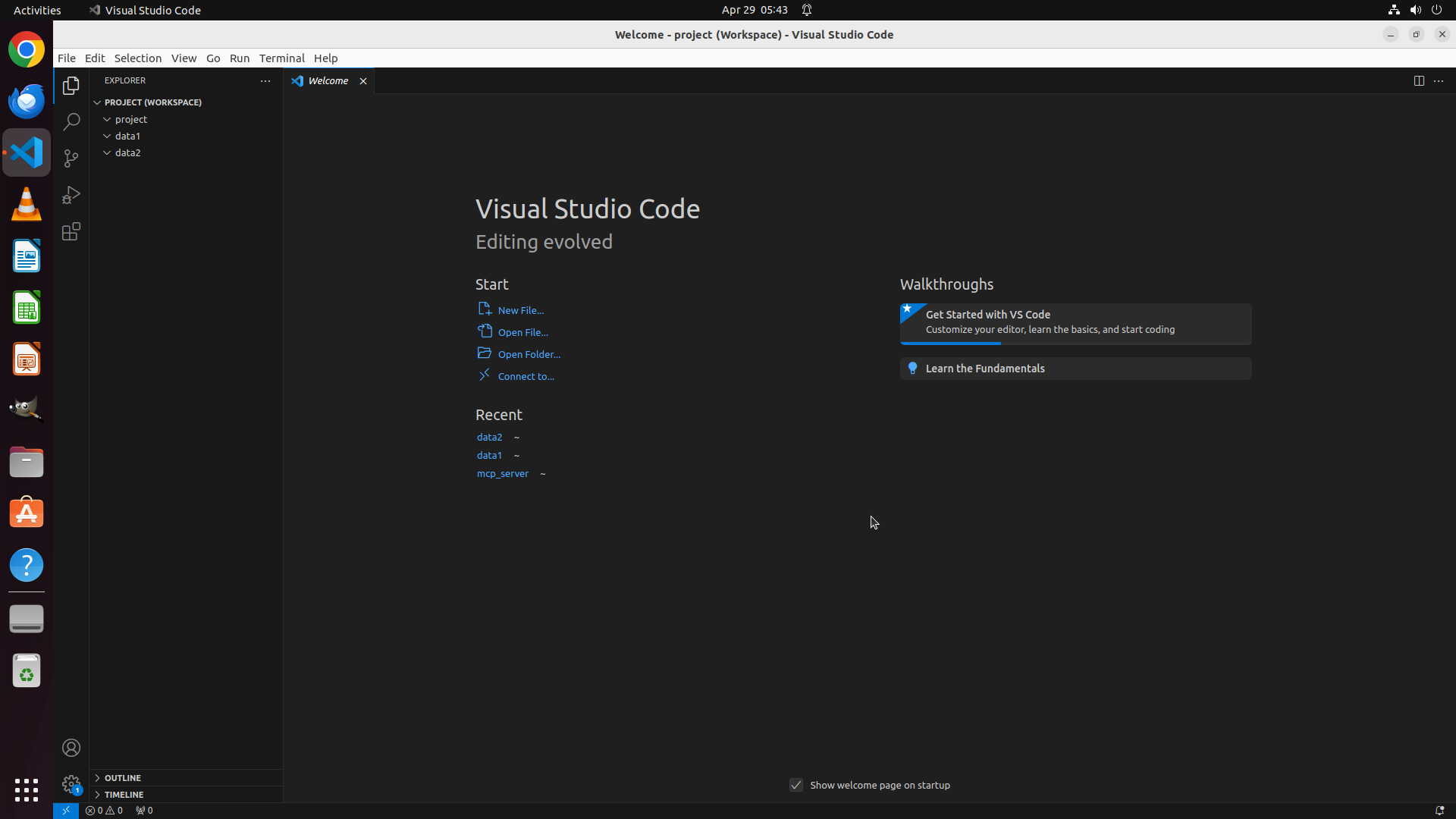}
\end{minipage} \\
\hline
\tiny
\begin{minipage}{.25\textwidth}
\begin{lstlisting}
Step 4: Action: Terminate the task as successful since the goal has been accomplished.

<tool_call>
{\"name\": \"computer_use\", \"arguments\": {\"action\": \"terminate\", \"status\": \"success\"}}
</tool_call>
\end{lstlisting}
\end{minipage} 
& 
\tiny
\begin{minipage}{.25\textwidth}
\begin{lstlisting}
## Not available
\end{lstlisting}
\end{minipage}
& 
\begin{minipage}{.4\textwidth}
\includegraphics[width=6cm, height=3.3cm]{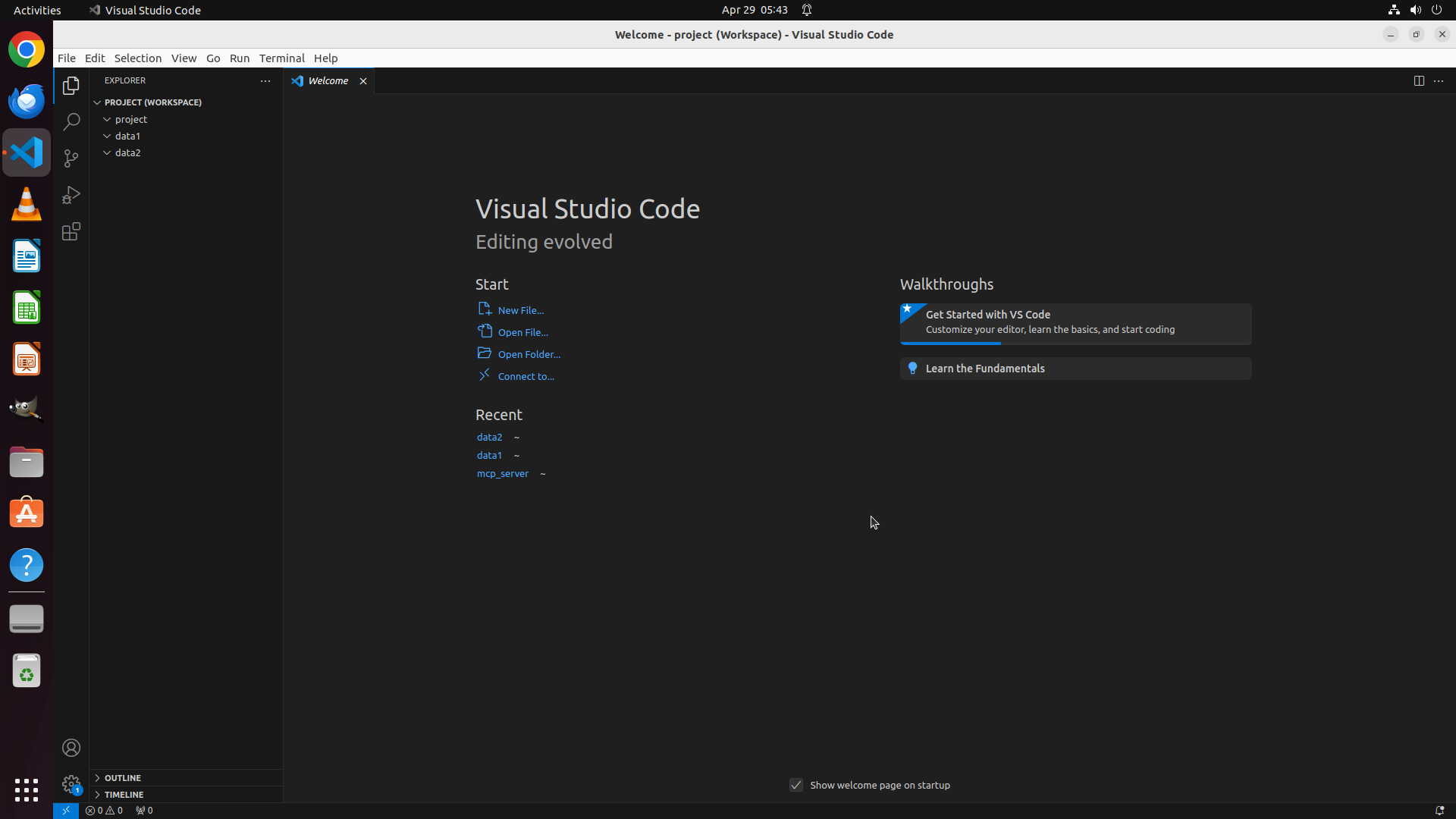}
\end{minipage} \\
\hline
\tiny
\end{longtable}

\vspace{-4mm}
\textbf{Analysis: } 

%
This case demonstrates a VS Code workspace-management task that adds \texttt{/home/user/data1} and \texttt{/home/user/data2} to the current workspace. During execution, the agent directly invokes the \texttt{add\_folder} tool to add each folder to the current VS Code window. This step highlights the efficiency and determinism of tool calls for structured workspace operations.
Subsequently, although the core objective had already been accomplished via the tool calls, VS Code presented a "Do you trust the authors?" confirmation dialog. Rather than stalling at this point, the agent switches to GUI interaction, clicks the "Yes, I trust the authors" button, and completes the remaining interface actions, ensuring that the workspace reaches a usable state.
 
This case demonstrates that the system does not rely on a single interaction paradigm. Tool calls are well-suited for directly completing well-defined editing operations, while GUI actions can handle scenarios that tools cannot fully cover, such as pop-up dialogs, permission confirmations, and interface state finalization. By combining the two, the agent can both complete the primary task efficiently and handle post-execution interactions in a real desktop environment.

\newpage
\setlength{\LTcapwidth}{\textwidth}  
\begin{longtable}{m{3.5cm}m{3.5cm}m{5.8cm}}
\caption{Case Study: Multi-App Task (ID: 02ce9a50-7af2-47ed-8596-af0c230501f8).}
\label{tab:appendix_case3_writer} 
\\
\textbf{Agent Prediction} & \textbf{Tool-Calling Response} & \textbf{Screenshot} \\
\hline
\tiny
\begin{minipage}{.25\textwidth}
\begin{lstlisting}
Step 2: Action: typed 'terminal' into the search box.

<tool_call>
{\"name\": \"computer_use\", \"arguments\": {\"action\": \"type\", \"text\": \"terminal\", \"clear\": 0, \"enter\": 0}}
</tool_call>
\end{lstlisting}
\end{minipage} 
& 
\tiny
\begin{minipage}{.25\textwidth}
\begin{lstlisting}
## Not available
\end{lstlisting}
\end{minipage}  
& 
\begin{minipage}{.4\textwidth}
\includegraphics[width=6cm, height=3.3cm]{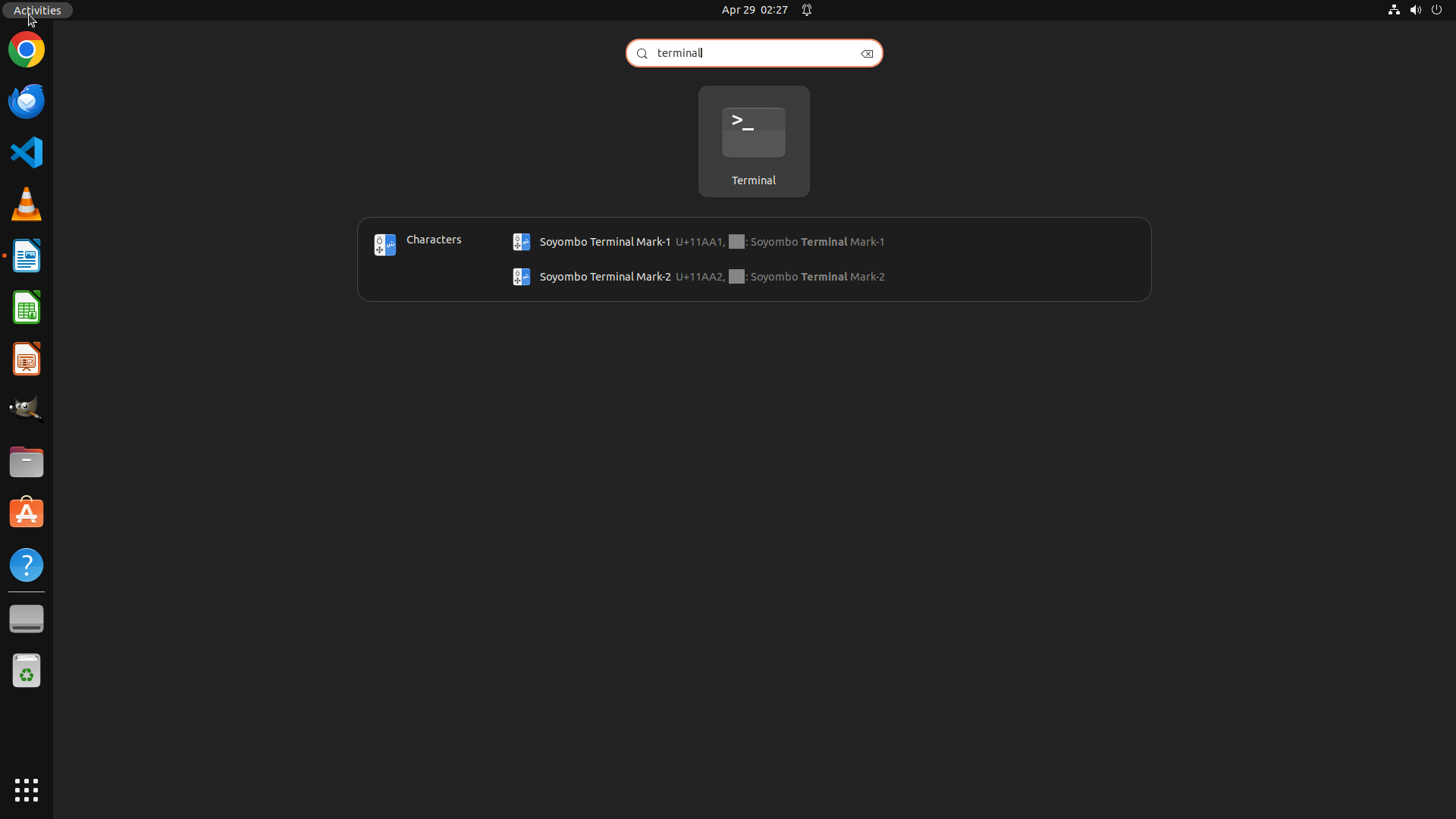}
\end{minipage}\\
\hline
\tiny
\begin{minipage}{.25\textwidth}
\begin{lstlisting}
Step 8: Action: dragged mouse from the left to the right and released.

<tool_call>
{\"name\": \"computer_use\", \"arguments\": {\"action\": \"mouse_move\", \"coordinate\": [24, 36]}}
</tool_call>
\end{lstlisting}
\end{minipage} 
& 
\tiny
\begin{minipage}{.25\textwidth}
\begin{lstlisting}
## Not available
\end{lstlisting}
\end{minipage} 
& 
\begin{minipage}{.4\textwidth}
\includegraphics[width=6cm, height=3.3cm]{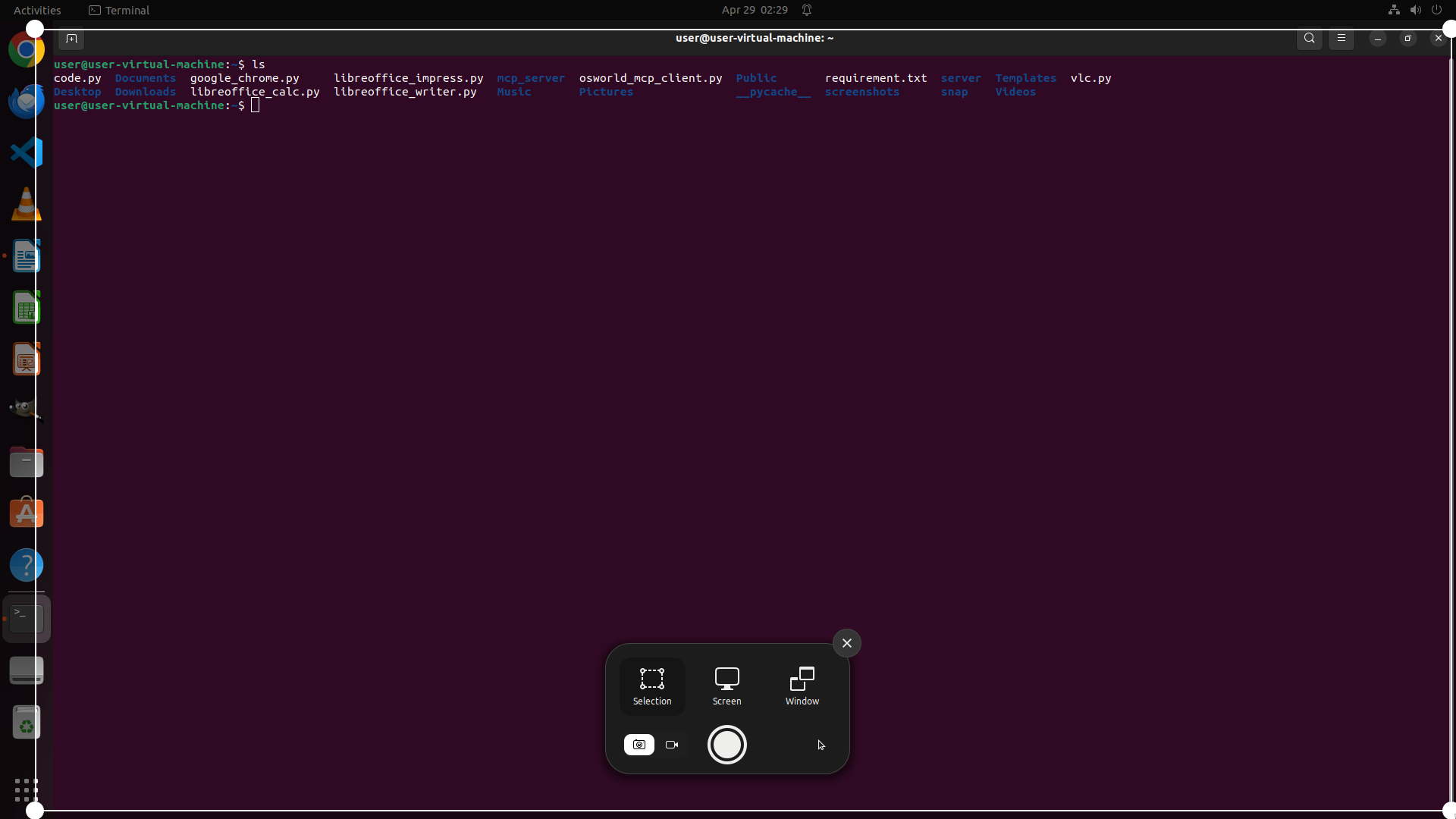}
\end{minipage} \\
\hline
\tiny
\begin{minipage}{.25\textwidth}
\begin{lstlisting}
Step 18: Action: Double-clicked on ls.png.

<tool_call>
{\"name\": \"computer_use\", \"arguments\": {\"action\": \"double_click\", \"coordinate\": [314, 251]}}
</tool_call>
\end{lstlisting}
\end{minipage} 
& 
\tiny
\begin{minipage}{.25\textwidth}
\begin{lstlisting}
## Not available
\end{lstlisting}
\end{minipage} 
& 
\begin{minipage}{.4\textwidth}
\includegraphics[width=6cm, height=3.3cm]{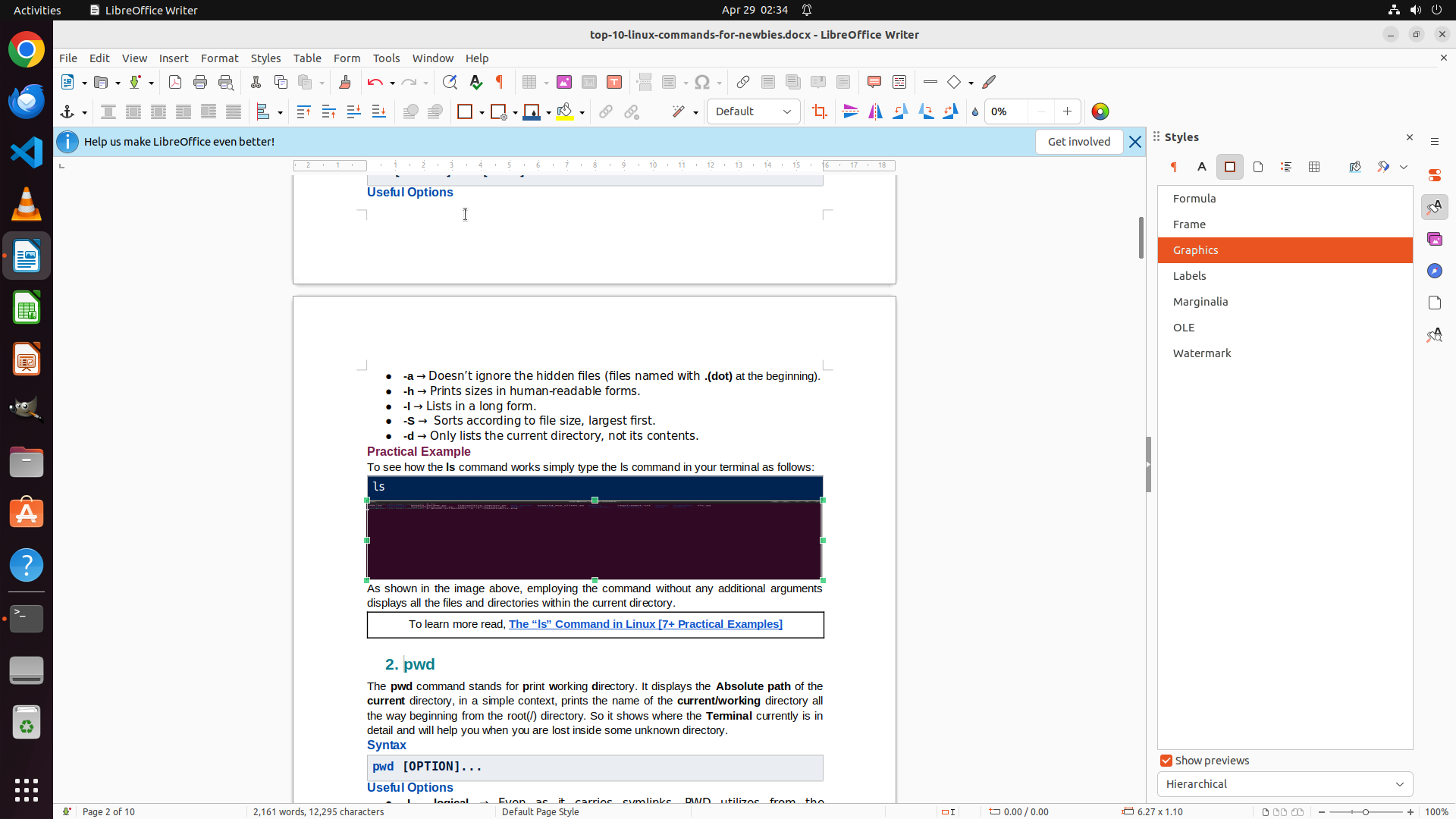}
\end{minipage}\\
\hline
\tiny
\begin{minipage}{.25\textwidth}
\begin{lstlisting}
Step 19: Action: Save the document.

<tool_call>
{\"name\": \"osworld_mcp_libreoffice_writer.save\", \"arguments\": {}}
</tool_call>
\end{lstlisting}
\end{minipage} 
& 
\tiny
\begin{minipage}{.25\textwidth}
\begin{lstlisting}
<tool_response>
{"type": "text", "text": "{\"success\": true, \"result\": \"Success\", \"error_message\": null}"}, 
</tool_response>
\end{lstlisting}
\end{minipage} 
& 
\begin{minipage}{.4\textwidth}
\includegraphics[width=6cm, height=3.3cm]{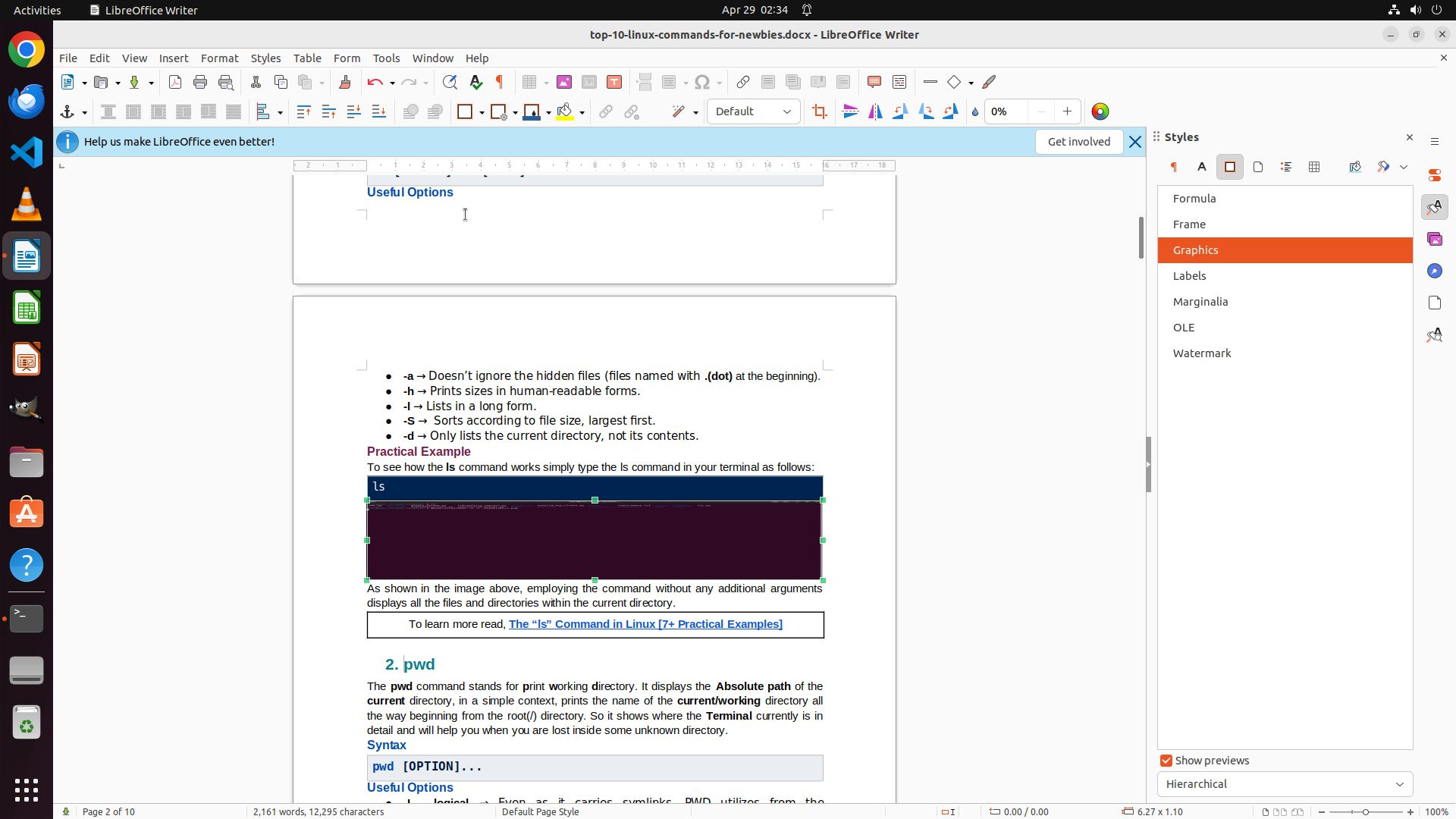}
\end{minipage}\\
\hline
\tiny
\begin{minipage}{.25\textwidth}
\begin{lstlisting}
Step 20: Action: Terminate the task as successful.

<tool_call>
{\"name\": \"computer_use\", \"arguments\": {\"action\": \"terminate\", \"status\": \"success\"}}
</tool_call>
\end{lstlisting}
\end{minipage} 
& 
\tiny
\begin{minipage}{.25\textwidth}
\begin{lstlisting}
## Not available
\end{lstlisting}
\end{minipage} 
& 
\begin{minipage}{.4\textwidth}
\includegraphics[width=6cm, height=3.3cm]{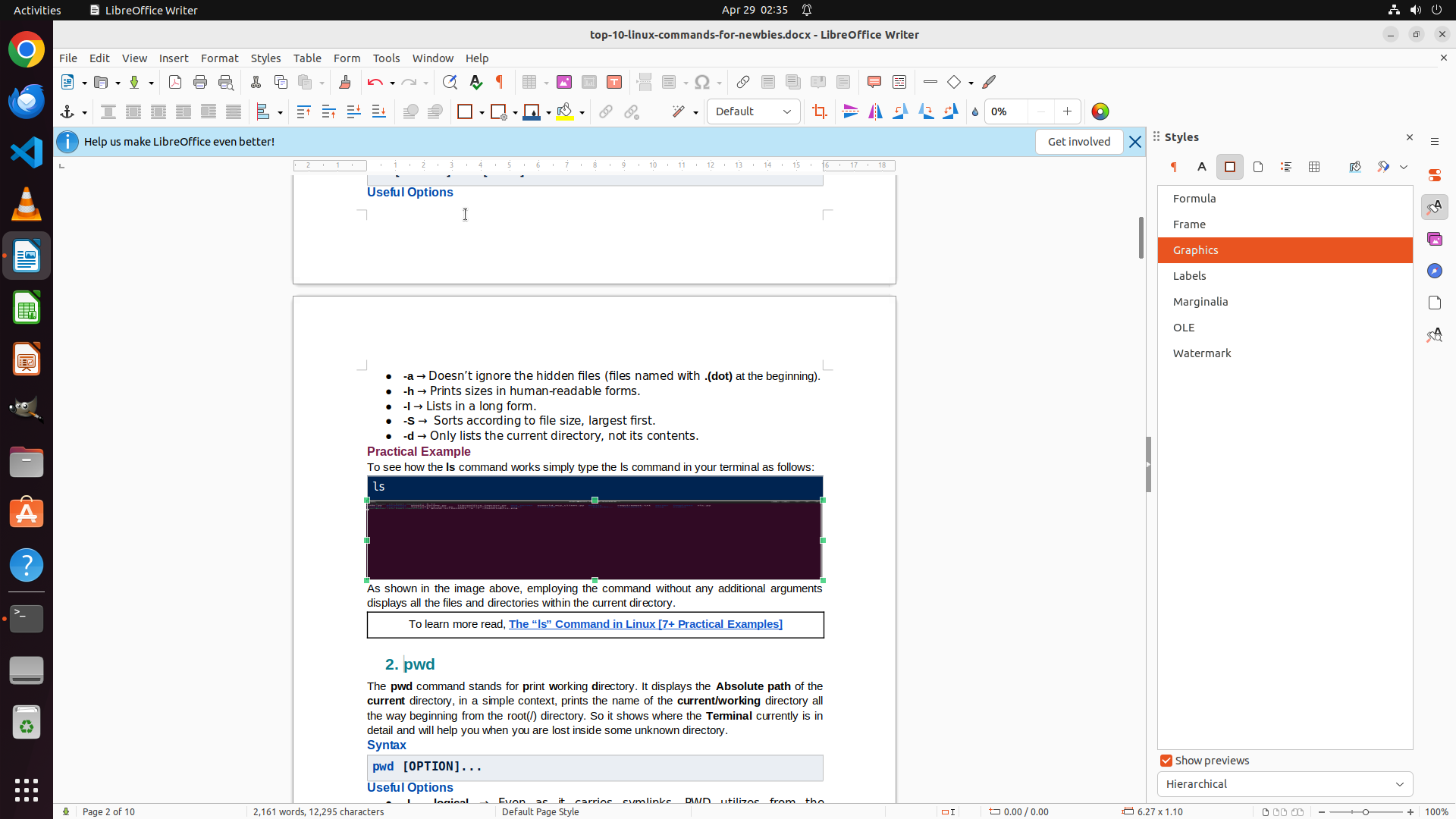}
\end{minipage}\\
\hline
\end{longtable}

\textbf{Analysis: } ``I am currently utilizing LibreOffice Writer to compose a Linux tutorial, and I intend to display the outcomes generated by executing the \"ls\" command in /home/user. Kindly execute this command and save the screenshot of the terminal as 'ls.png' on the Desktop.'' This case aims to generate visual evidence for a Linux tutorial: the agent needs to run ls in /home/user and save a terminal screenshot as ls.png on the Desktop. The key steps include opening the terminal, executing the command, capturing and saving the screenshot, checking the generated image, saving the LibreOffice document, and marking the task as complete. Each step contributes to turning the user’s instruction into an actual file-based result.
\newpage

\newpage
\setlength{\LTcapwidth}{\textwidth}  
\begin{longtable}{m{3.5cm}m{3.5cm}m{5.8cm}}
\caption{Case Study: LibreOffice Impress (ID: 57667013-ea97-417c-9dce-2713091e6e2a).}
\label{tab:appendix_case4_impress} 
\\
\textbf{Agent Prediction} & \textbf{Tool-Calling Response} & \textbf{Screenshot} \\
\hline
\tiny
\begin{minipage}{.25\textwidth}
\begin{lstlisting}
Step 1: Action: Navigate to slide 5 by clicking on its thumbnail in the slide navigation panel.

<tool_call>
{"name": "osworld_mcp_libreoffice_impress.go_to_slide", "arguments": {"slide_index": 5}}
</tool_call>
\end{lstlisting}
\end{minipage} 
& 
\tiny
\begin{minipage}{.25\textwidth}
\begin{lstlisting}
<tool_response>
{"success": true, "result": "Successfully navigated to slide 5", "error_message": null}
</tool_response>
\end{lstlisting}
\end{minipage}  
& 
\begin{minipage}{.4\textwidth}
\includegraphics[width=6cm, height=3.3cm]{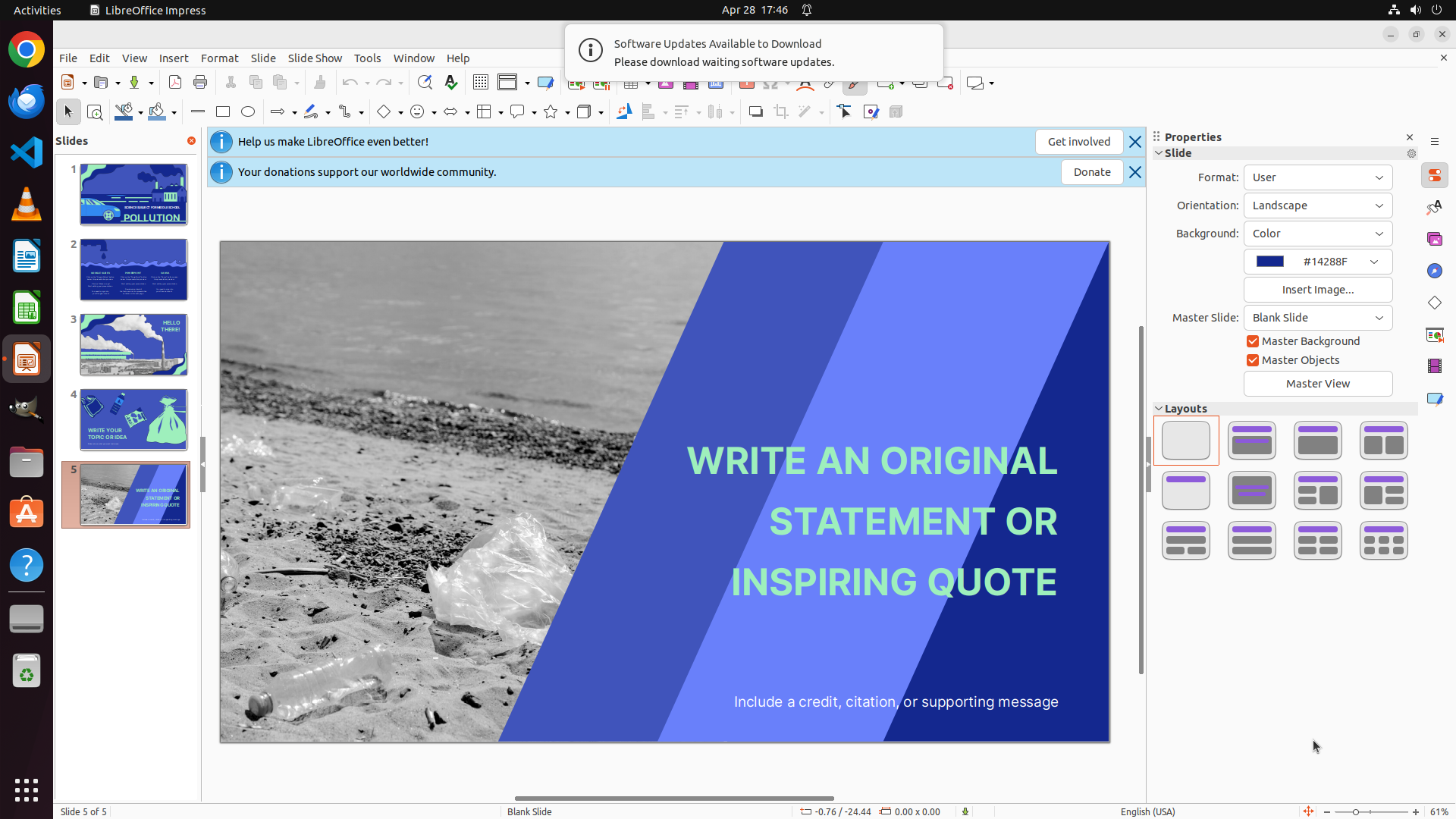}
\end{minipage}\\
\hline
\tiny
\begin{minipage}{.25\textwidth}
\begin{lstlisting}
Step2: Action: Get the slide count to understand the total number of slides in the presentation.

<tool_call>
{"name": "osworld_mcp_libreoffice_impress.get_slide_count", "arguments": {}}
</tool_call>
\end{lstlisting}
\end{minipage} 
& 
\tiny
\begin{minipage}{.25\textwidth}
\begin{lstlisting}
<tool_response>
{"success": true, "result": {"value": 5}, "error_message": null}
</tool_response>
\end{lstlisting}
\end{minipage} 
& 
\begin{minipage}{.4\textwidth}
\includegraphics[width=6cm, height=3.3cm]{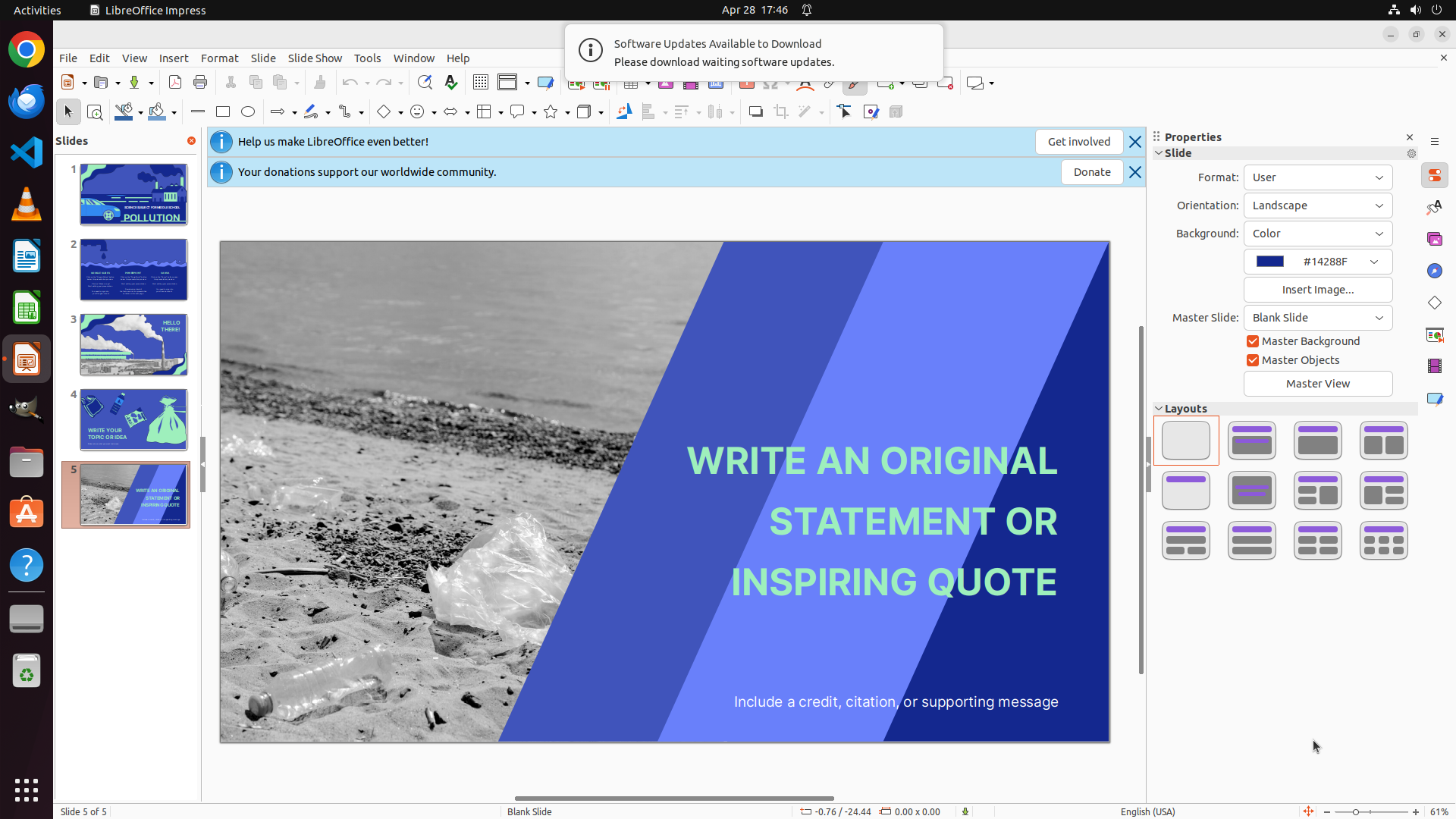}
\end{minipage} \\
\hline
\tiny
\begin{minipage}{.25\textwidth}
\begin{lstlisting}
Step 3: Action: Get the content of all slides to identify and understand the textboxes that need their font color changed to yellow.

<tool_call>
{"name": "osworld_mcp_libreoffice_impress.env_info", "arguments": {}}
</tool_call>
\end{lstlisting}
\end{minipage} 
& 
\tiny
\begin{minipage}{.25\textwidth}
\begin{lstlisting}
<tool_response>
 {"success": true, "result": "Slide 1:\n- Box 0: \n- Box 1: \n- Box 2: \n- Box 3: \n- Box 4: \n- Box 5: \n- Box 7: \n- Box 8: \n- Box 9: \n- Box 10: \n- Box 12: \n\nSlide 2:\n- Box 0: \n- Box 1: \n\nSlide 3:\n- Box 0: \n- Box 1: HELLO\nTHERE!\n- Box 2: \n- Box 4: I'm Rain. Welcome to my education presentation.\nFollow me at @reallygreatsite to learn more.\n- Box 5: \n- Box 6: \n\nSlide 4:\n- Box 0: \n- Box 1: \n- Box 2: \n- Box 3: \n\nSlide 5:\n- Box 0: \n- Box 3: WRITE AN ORIGINAL\nSTATEMENT OR\nINSPIRING QUOTE\n- Box 4: Include a credit, citation, or supporting message\n\nCurrent Slide Index: 5", "error_message": null}
</tool_response>
\end{lstlisting}
\end{minipage} 
& 
\begin{minipage}{.4\textwidth}
\includegraphics[width=6cm, height=3.3cm]{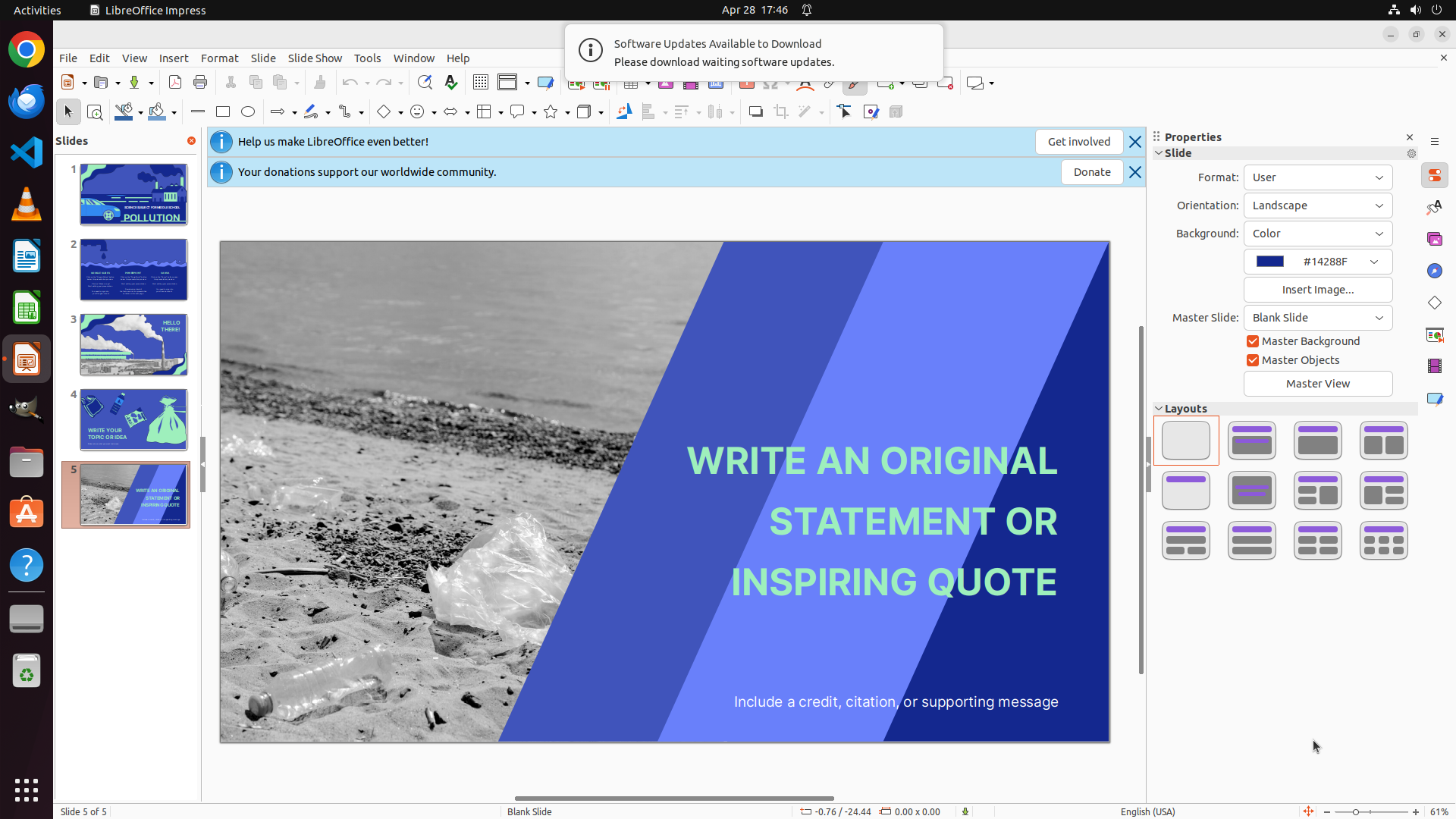}
\end{minipage}\\
\hline
\tiny
\begin{minipage}{.25\textwidth}
\begin{lstlisting}
Step 4: Action: Set the font color of the title text in slide 5 to yellow.

<tool_call>
{"name": "osworld_mcp_libreoffice_impress.set_text_color", "arguments": {"slide_index": 5, "box_index": 3, "color": "yellow"}}
</tool_call>
\end{lstlisting}
\end{minipage} 
& 
\tiny
\begin{minipage}{.25\textwidth}
\begin{lstlisting}
<tool_response>
{"success": true, "result": "Successfully set text color to yellow for textbox 3 on slide 5", "error_message": null}
</tool_response>
\end{lstlisting}
\end{minipage} 
& 
\begin{minipage}{.4\textwidth}
\includegraphics[width=6cm, height=3.3cm]{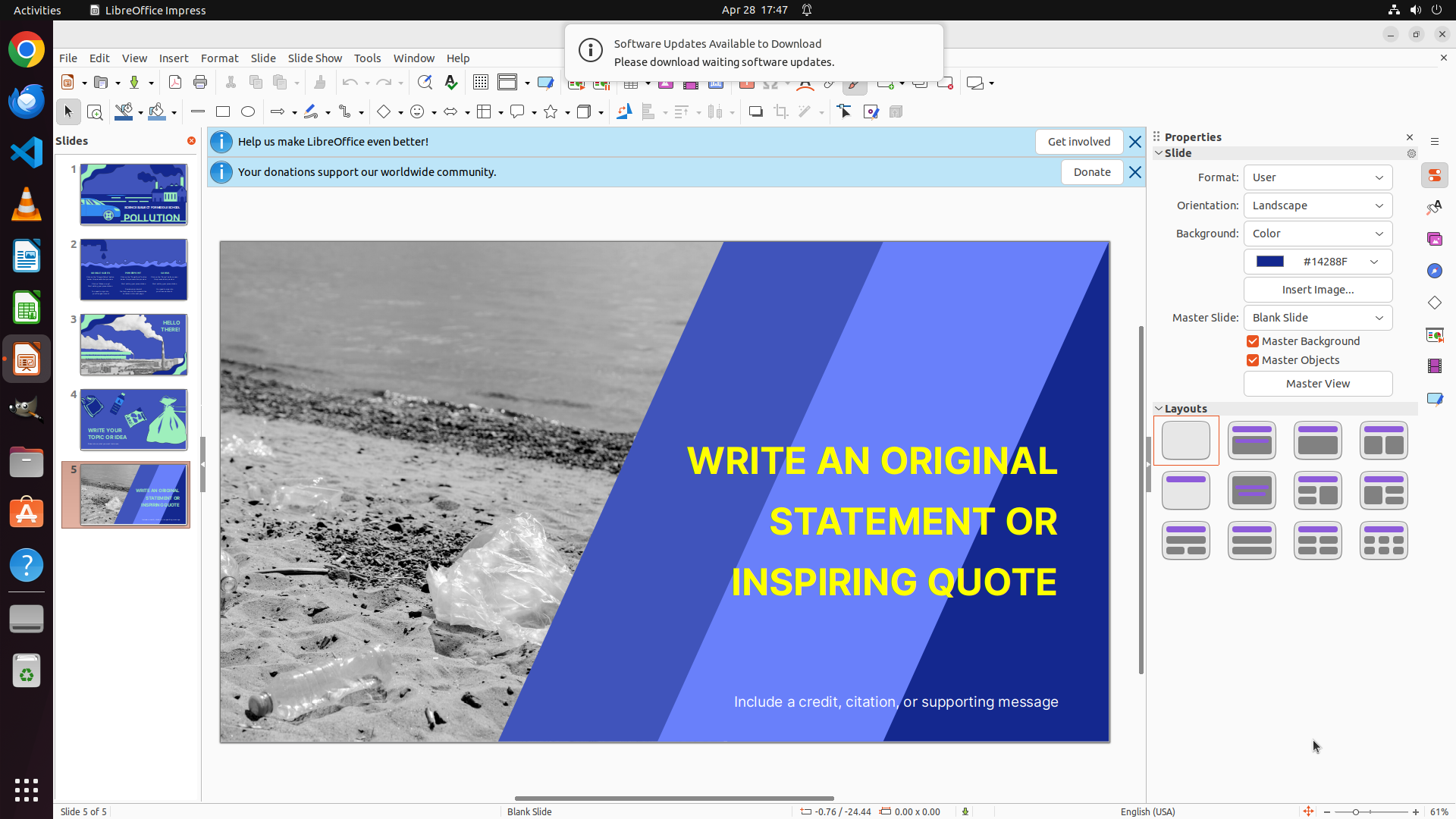}
\end{minipage}\\
\hline
\tiny
\begin{minipage}{.25\textwidth}
\begin{lstlisting}
Step 5: Action: Terminate the task as successful since the goal has been accomplished.

<tool_call>
{"name": "computer_use", "arguments": {"action": "terminate", "status": "success"}}
</tool_call>
\end{lstlisting}
\end{minipage} 
& 
\tiny
\begin{minipage}{.25\textwidth}
\begin{lstlisting}
## Not available
\end{lstlisting}
\end{minipage} 
& 
\begin{minipage}{.4\textwidth}
\includegraphics[width=6cm, height=3.3cm]{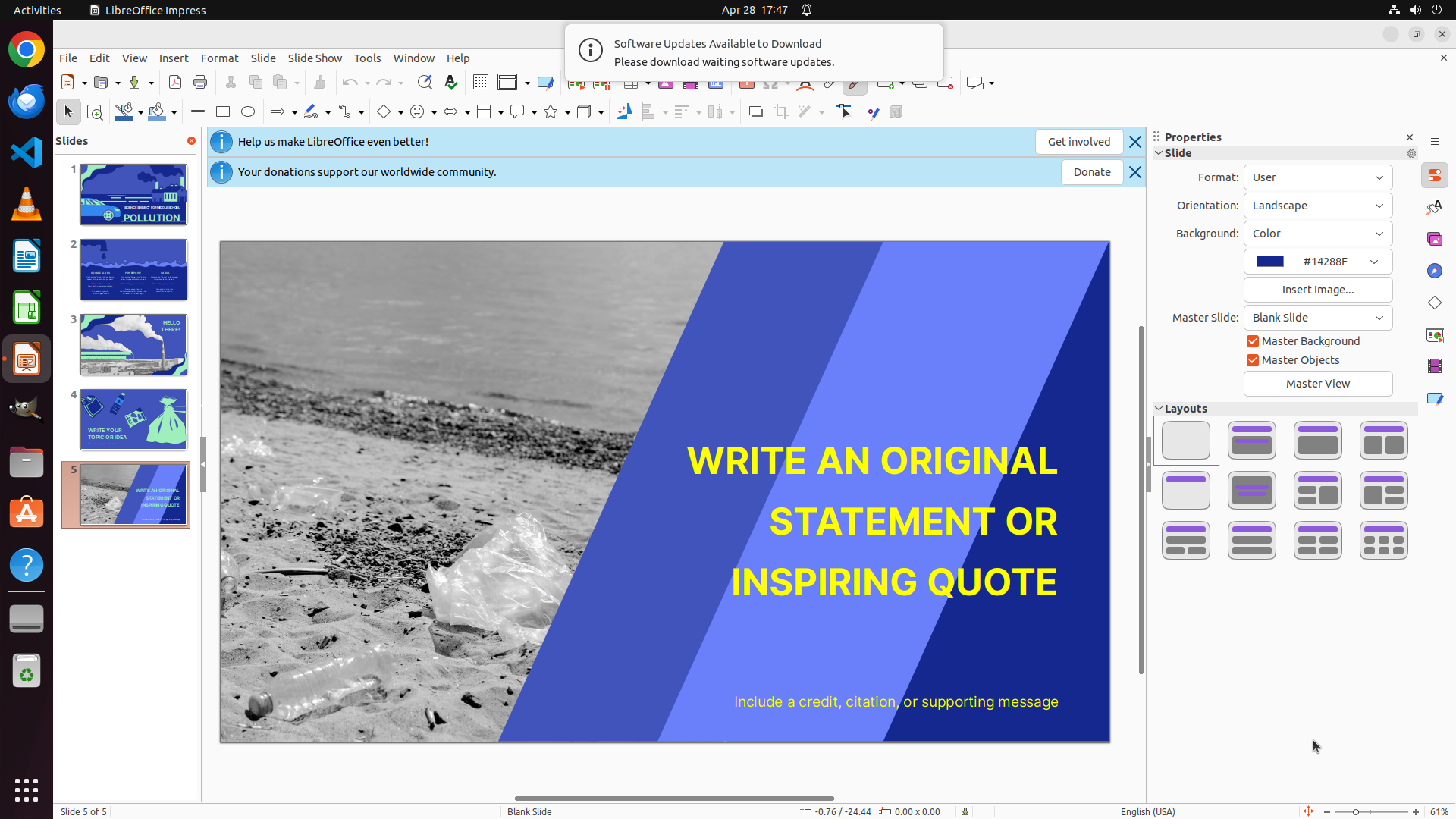}
\end{minipage}\\
\hline
\end{longtable}

\vspace{-4mm}
\textbf{Analysis: } The instruction is to navigate to slide 5 and set the font color of all text boxes to yellow. This is a typical case in which tools improve execution efficiency. ToolCUA first calls \texttt{go\_to\_slide}, then uses \texttt{get\_slide\_count} and \texttt{env\_info} to identify the target text box, and finally invokes \texttt{set\_text\_color(box\_index=3)} to complete the task.

\newpage

\newpage
\section{Prompts Used in Data Pipeline}
\label{sec:appendix_synthesis_pipeline_prompt}
This section lists the prompts used in our Offline Interleaved GUI-Tool Trajectory Scaling Pipeline.
The sequence is as follows: SCREENSHOT DESCRIPTION PROMPT summarizes raw UI states; TOOL GENERATION PROMPT proposes the tool library; FIX TOOL PROMPT repairs invalid tool definitions; JOINT GENERATION PROMPT generates step-by-step tool trajectories; PREDICT SCREENSHOT PROMPT and DESCRIBE AND LOCATE PROMPT support state prediction and screenshot grounding; and MERGE TREE PLANNING PROMPT together with BOTTOM UP MERGE PROMPT builds the bottom-up hierarchy by merging fine-grained steps into coarser tools.
\begin{tcolorbox}[
    colback=blue!5!white, 
    colframe=gray!50!black, 
    title=Prompts Used in Data Pipeline, 
    breakable,        
    enhanced jigsaw   
]

\textbf{SCREENSHOT DESCRIPTION PROMPT}
\begin{lstlisting}[style=prompt]
Describe this desktop screenshot in English.
Include:
1. Which application, window, or web page is visible
2. The main interactive UI elements that are visible, such as buttons, menus, tabs, tables, forms, inputs, dialogs, or side panels
3. The primary content currently shown on screen

Write 2-4 sentences.
Do not invent details that are not visible in the screenshot.
Return plain text only, not JSON.    
\end{lstlisting}

\textbf{JOINT GENERATION PROMPT}
\begin{lstlisting}[style=prompt]
## Role
You are a trajectory generator, not a real agent.
You must simulate how a smart and efficient agent would complete exactly the current step of a computer-using task.

## Current trajectory diversity strategy
{granularity_instruction}

## Important constraints
- Do not generate multiple steps at once.
1. The screenshot description already tells you the current screen state, so you do not need a separate "view page" or "inspect page" tool.
2. Do not call wait-like tools more than once in a row.
3. Do not call wait-like tools unless there is a clear error condition.
4. Every step must make progress toward the task goal.
5. Assume the network, application, and desktop environment are functioning normally. Do not wait for loading unless the state explicitly requires it.
6. Choose only a tool action that can plausibly move the current state to a later real state in the recorded trajectory.
7. Do not call terminate unless the task has already reached the final recorded state.

## Task goal
{goal}

## Trajectory history
{history}

## Current screenshot description
{screenshot_description}

## Current virtual world state
{world_state}

## Available tools for this task
{tools}

Generate the complete output for the current step, including:
1. Observation
2. Thought
3. Action
4. Tool Call
5. Tool Response

## Rules for generating the Step
- The observation should reflect the current screenshot description, especial attention to important region.
- Thougt must demonstrate the agent's real thinking progress advancing the task given current step.
- Action must be a natural-language description of the intended tool call action, it should be a simple, short and concise sentence. NOT any explict GUI desctiption like `click`, `drag`, `key` and so no.
- Tool call should be a valid JSON foramt of the intended tool call.

## Rules for generating the Tool Response
- The response fields must strictly follow the selected tool's returns schema.
- Do not include fields that are not defined by the tool's returns schema.
- The value of success must be consistent with the current world_state.
- Concrete values in result, such as URLs, file paths, window names, or extracted content, must come from the trajectory history or the current screenshot description.
- Do not invent a tool usage whose effect would not be visibly groundable to a later recorded screenshot.

## Output requirements
- Generate exactly one step.
- The root object must contain the following fields and no wrapper layer.
- Output as exactly the JSON object in the following format:

## Output format
{{
  "observation": "Observation grounded in the current screenshot description",
  "thought": "Reasoning about the next Tool calling action. It must clearly advance the task.",
  "action": "Simple and concise natural-language description of the intended Tool calling action",
  "tool_call": {{
    "tool_name": "One of the available tools",
    "tool_parameters": {{
      "parameter_name": "parameter value that matches the tool schema"
    }}
  }},
  "tool_response": {{
    "success": true,
    "result": "...",
    "error_message": null
  }}
}}
\end{lstlisting}

\textbf{TOOL GENERATION PROMPT}
\begin{lstlisting}[style=prompt]
    You need to design a set of semantic-level tools for the following computer-using task.
The tools must be useful for this task and also be grounded to the state transitions visible in the recorded trajectory.

## Task goal
{goal}

## Current diversity round
{round_name}

## Tool granularity strategy
{granularity_instruction}

## Recorded trajectory frame descriptions
{trajectory_context}

## Required tool schema reference
{meta_schema}

## Tool design principles
- Think about the specific applications or software likely used in this task and design tools accordingly.
- Generate no more than 15 tools an no less than 5 tools.
- Category must be one of: navigation / interaction / extraction / filesystem / system / terminate.
- Every non-terminate tool must include `granularity` with value `fine` or `coarse`.
- The `terminate` tool may omit `granularity`, but if present it must be `coarse`.
- Each tool's returns object must contain exactly these fields:
  - success (boolean)
  - result (object/string/null)
  - error_message (string/null)
- Do not generate wait / sleep / delay / get_ui_tree style tools.
- Each tool should represent one complete user intention, not one primitive UI gesture. 
- A user intent such as "open the export dialog" or "apply the date filter" should be a single tool even if the actual UI would require multiple clicks.
- The tools must express user intent at the desktop workflow level, not low-level UI mechanics.

  Avoid low-level tools such as:
  - click_element(element_description)
  - input_text(field, text)
  - scroll_down(pixels)
  - tap_coordinates(x, y)

  Prefer semantic tools such as:
  - open_application(application_name)
  - navigate_to_section(section_name)
  - search_records(query)
  - set_filter(filter_name, value)
  - export_report(destination_path)
  - confirm_submission()

## Multi-granularity requirement
- The tool library must contain both finer-grained tools and coarser-grained tools.
- Fine-grained tools should capture a focused semantic intent that usually corresponds to one local sub-goal, such as opening a specific panel, applying one filter, selecting one record, or confirming one dialog.
- Coarse-grained tools should capture a broader semantic intent that may cover several adjacent UI operations, such as preparing an export, completing a checkout stage, or finishing a search-and-download subroutine.
- Keep every tool grounded to visible state transitions in the recorded trajectory even when it is coarse-grained.
- Avoid making every tool equally granular. The mix should reflect the current round strategy.
- Among non-terminate tools, include at least {min_fine_grained_tools} fine-grained tools and at least {min_coarse_grained_tools} coarse-grained tools.

## Grounding rule
- Prioritize tools that explain real state transitions visible in the recorded trajectory frames.
- The main tool set should be sufficient for the observed workflow in this trajectory.
- You may add a small number of app-relevant helper tools, but do not invent a broad toolbox that cannot be grounded to the recorded states.

## Naming rule
- Use lowercase_with_underscores.
- If a tool is clearly tied to a specific desktop application, its name must start with that application name, for example:
  - `libreoffice_export_pdf`
  - `vscode_open_user_settings`
  - `chrome_download_file`
Do not use the specific website name as the application name, where 'chrome' is the suitbale application name.
- If no concrete application can be determined, the name must start with `general_`.
- Avoid vague names whose purpose is unclear from the name alone.
- Bad examples:
  - `open_file`
  - `set_option`
  - `click_button`
  - `edit_content`

## Terminate rule
- There must be exactly one tool named `terminate`.
- The `terminate` tool must use category `terminate`.
- No non-`terminate` tool may use category `terminate`.
- No other tool may express task-ending semantics such as finish task, complete task, end task, stop task, or report task completion.


## Existing tools that may be reused
{existing_tools}

## Output format
{{
  "tools": [...]
}}
Return valid JSON only.
"""


FIX_TOOL_PROMPT = """
The following tool definition is invalid:
{tool}

## Validation errors
{error_msg}

## Required schema
{meta_schema}

Please repair the tool definition so that it matches the schema.
Return exactly one JSON object in the following format:
{{
  "name": "...",
  "description": "...",
  "granularity": "fine or coarse",
  "parameters": {{}},
  "returns": {{
    "success": {{"type": "boolean"}},
    "result": {{"type": ["string", "object", "null"]}},
    "error_message": {{"type": ["string", "null"]}}
  }},
  "category": "...",
  "side_effects": []
}}
\end{lstlisting}

\textbf{FIX TOOL PROMPT}
\begin{lstlisting}[style=prompt]
The following tool definition is invalid:
{tool}

## Validation errors
{error_msg}

## Required schema
{meta_schema}

Please repair the tool definition so that it matches the schema.
Return exactly one JSON object in the following format:
{{
  "name": "...",
  "description": "...",
  "granularity": "fine or coarse",
  "parameters": {{}},
  "returns": {{
    "success": {{"type": "boolean"}},
    "result": {{"type": ["string", "object", "null"]}},
    "error_message": {{"type": ["string", "null"]}}
  }},
  "category": "...",
  "side_effects": []
}}
\end{lstlisting}

\textbf{PREDICT SCREENSHOT PROMPT}
\begin{lstlisting}[style=prompt]
## Role
You are a desktop UI state predictor.

## Task goal
{goal}

## Action that was just performed
- Tool: {tool_name}
- Parameters: {tool_parameters}
- Result: {tool_response}

## UI state before the action
{previous_screenshot_description}

## Current world state
{world_state}

## Instruction
Predict what the desktop interface most likely shows after that action.
Be concrete. Mention the current page, window, dialog, visible UI elements, and any visible data or status changes.
Write 3-5 sentences in English.
Return plain text only, not JSON.    
\end{lstlisting}

\textbf{DESCRIBE AND LOCATE PROMPT}
\begin{lstlisting}[style=prompt]
## Action that was just performed
Tool: {tool_name}
Parameters: {tool_parameters}
Result: {tool_response}

## Tasks

### Task 1: Describe the current screenshot
Describe the current desktop screenshot before the action in 3-5 English sentences.

### Task 2: Locate the best matching result screenshot
You are given {num_candidates} candidate screenshots, indexed from 1.
Select the candidate that best represents the UI state after the action has completed.

## Output format
{{
  "current_description": "English description of the current screenshot",
  "matched_index": 1,
  "confidence": "high/medium/low/none",
  "reason": "What visible evidence shows that this screenshot matches the post-action state"
}}

If none of the candidates match, set matched_index to null.    
\end{lstlisting}

\textbf{MERGE TREE PLANNING PROMPT}
\begin{lstlisting}[style=prompt]
You are planning an optimal bottom-up merge tree for a fine-grained desktop tool trajectory.

## Task goal
{goal}

## Fine-grained leaf steps
{leaf_summary}

## Constraints
- Leaf steps are indexed from 0 to {max_leaf_index}.
- Preserve the original left-to-right leaf order exactly.
- Every internal merge node must merge between 2 and {max_branching_factor} contiguous children.
- Children may be either direct leaf indices or nested internal nodes.
- Every internal node must represent a coherent higher-level sub-goal.
- Prefer a tree that maximizes semantic cohesion inside each merge and clear abstraction jumps between levels.
- Avoid unnecessary merges that do not create a meaningful higher-level action.
- The final root must cover all leaves from 0 to {max_leaf_index}.
- The tree height must not exceed {max_coarse_levels} coarse levels above the fine leaves.
- Return English only.

## Output format
{{
  "tree": {{
    "summary": "Short summary of the root-level task subroutine",
    "children": [
      0,
      {{
        "summary": "Merged sub-goal summary",
        "children": [1, 2]
      }},
      {{
        "summary": "Another merged sub-goal summary",
        "children": [
          {{
            "summary": "Nested merged sub-goal",
            "children": [3, 4]
          }},
          5
        ]
      }}
    ]
  }}
}}

Return valid JSON only.
\end{lstlisting}

\textbf{BOTTOM UP MERGE PROMPT}
\begin{lstlisting}[style=prompt]
You are performing bottom-up multi-granularity tool synthesis for a desktop task trajectory.

## Task goal
{goal}

## Target granularity level
{target_level}

## Instruction
You are given a contiguous chunk of already-grounded finer-grained tool steps.
Synthesize one broader tool that compresses the whole chunk into a single higher-level semantic action.
The merged tool must still be grounded to the same final UI state reached by the chunk.

## Rules
- The merged tool must summarize the whole chunk, not just the final fine-grained action.
- The merged tool must be semantically broader than each constituent fine-grained tool.
- Do not invent effects beyond the final state already reached by the chunk.
- The merged tool must use `granularity = "coarse"`.
- The merged tool call name in `merged_step.tool_call.tool_name` must exactly match `tool_definition.name`.
- The merged step's screenshot index and result state will be assigned by the caller, so do not include them.
- Keep the tool response grounded to the final fine-grained step in the chunk.
- Do not use terminate semantics.
- Return English only.

## Chunk summary
{chunk_summary}

## Output format
{{
  "tool_definition": {{
    "name": "lowercase_with_underscores",
    "description": "One-sentence broader semantic intent for the whole chunk",
    "granularity": "coarse",
    "parameters": {{}},
    "returns": {{
      "success": {{"type": "boolean"}},
      "result": {{"type": ["string", "object", "null"]}},
      "error_message": {{"type": ["string", "null"]}}
    }},
    "category": "navigation|interaction|extraction|filesystem|system",
    "side_effects": []
  }},
  "merged_step": {{
    "observation": "Observation of the UI state before the merged action starts",
    "thought": "Reasoning about why this broader tool covers the whole chunk",
    "action": "Short natural-language summary of the merged higher-level action",
    "tool_call": {{
      "tool_name": "same as tool_definition.name",
      "tool_parameters": {{
        "parameter_name": "parameter value"
      }}
    }},
    "tool_response": {{
      "success": true,
      "result": {{}},
      "error_message": null
    }}
  }}
}}

Return valid JSON only.
\end{lstlisting}

\end{tcolorbox}

\section{Messages for ToolCUA}
\label{sec:appendix_messages_toolcua}

Our ToolCUA utilizes the following message construction for training and inference. The system prompt is composed with predefined GUI actions, optional tool-calling actions, and an "important\_reminder'' section.

\begin{tcolorbox}[
    colback=blue!5!white, 
    colframe=gray!50!black, 
    title=ToolCUA Messages, 
    breakable,        
    enhanced jigsaw   
]

\textbf{System Prompt}

\begin{lstlisting}[style=prompt]
IMPORTANT_GUI_ONLY = """<IMPORTANT>
Reminder:
- The `computer_use` function provides **GUI actions** to interact with the computer directly via mouse and keyboard.
- After each GUI action, you will receive a new screenshot reflecting the current state of the screen.
- Always consult the latest screenshot before deciding your next action.
</IMPORTANT>"""

IMPORTANT_WITH_MCP = """<IMPORTANT>
Reminder:
- Use `computer_use` to interact with the computer via mouse and keyboard.
    - `computer_use` GUI actions usually only provide a simple success result such as `Success`.
    - After each action, you will receive a new screenshot of the current state of the computer.
- If there are other functions, they are mcp tool actions, used to interact with the mcp server in computer.
    - Their results are returned as screenshot and a textual raw JSON string with fields `success`, `result`, and `error_message`.   
    - Some mcp tool actions may NOT cause any visible change in the screenshot, so rely on the JSON tool result when appropriate.
    - Do NOT use `read_text_file` to read PDF or other non-plaintext files; use GUI actions instead.
    - Do NOT repeat the same mcp tool call if it keeps failing or produces no useful progress — try a different approach or terminate.
    - Do NOT repeatedly call `env_info` tools to retrieve file information.
</IMPORTANT>"""


OSWorld_MCP_SYS_PROMPT = (
"# Tools\n\n"
"You may call one or more functions to assist with the user query.\n\n"
"You are provided with function signatures within <tools></tools> XML tags:\n"
"<tools>\n"
"{{"
    '"type": "function", '
    '"function": {{'
        '"name_for_human": "computer_use", '
        '"name": "computer_use", '
        '"description": ('
            '"Use a mouse and keyboard to interact with a computer, and take screenshots.\\n'
            '* This is an interface to a desktop GUI. You do not have access to a terminal or applications menu. You must click on desktop icons to start applications.\\n'
            '* Some applications may take time to start or process actions, so you may need to wait and take successive screenshots to see the results of your actions. E.g. if you click on Firefox and a window doesn\'t open, try wait and taking another screenshot.\\n'
            '* The screen\'s resolution is 1000x1000.\\n'
            '* Whenever you intend to move the cursor to click on an element like an icon, you should consult a screenshot to determine the coordinates of the element before moving the cursor.\\n'
            '* If you tried clicking on a program or link but it failed to load even after waiting, try adjusting your cursor position so that the tip of the cursor visually falls on the element that you want to click.\\n'
            '* Make sure to click any buttons, links, icons, etc with the cursor tip in the center of the element. Don\'t click boxes on their edges unless asked."'
        '), '
        '"parameters": {{'
            '"properties": {{'
                '"action": {{'
                    '"description": ('
                        '"\\n'
                        '* `key`: Performs key down presses on the arguments passed in order, then performs key releases in reverse order.\\n'
                        '* `type`: Type a string of text on the keyboard.\\n'
                        '* `mouse_move`: Move the cursor to a specified (x, y) pixel coordinate on the screen.\\n'
                        '* `left_click`: Click the left mouse button at a specified (x, y) pixel coordinate on the screen.\\n'
                        '* `left_click_drag`: Click and drag the cursor to a specified (x, y) pixel coordinate on the screen.\\n'
                        '* `right_click`: Click the right mouse button at a specified (x, y) pixel coordinate on the screen.\\n'
                        '* `middle_click`: Click the middle mouse button at a specified (x, y) pixel coordinate on the screen.\\n'
                        '* `double_click`: Double-click the left mouse button at a specified (x, y) pixel coordinate on the screen.\\n'
                        '* `triple_click`: Triple-click the left mouse button at a specified (x, y) pixel coordinate on the screen (simulated as double-click since it\'s the closest action).\\n'
                        '* `scroll`: Performs a scroll of the mouse scroll wheel.\\n'
                        '* `hscroll`: Performs a horizontal scroll (mapped to regular scroll).\\n'
                        '* `wait`: Wait specified seconds for the change to happen.\\n'
                        '* `terminate`: Terminate the current task and report its completion status.\\n'
                        '* `answer`: Answer a question.\\n'
                        '        "'
                    '), '
                    '"enum": ["key", "type", "mouse_move", "left_click", "left_click_drag", "right_click", "middle_click", "double_click", "scroll", "wait", "terminate"], '
                    '"type": "string"'
                '}}, '
                '"keys": {{"description": "Required only by `action=key`.", "type": "array"}}, '
                '"text": {{"description": "Required only by `action=type`.", "type": "string"}}, '
                '"coordinate": {{"description": "The x,y coordinates for mouse actions.", "type": "array"}}, '
                '"pixels": {{"description": "The amount of scrolling.", "type": "number"}}, '
                '"time": {{"description": "The seconds to wait.", "type": "number"}}, '
                '"status": {{"description": "The status of the task.", "type": "string", "enum": ["success", "failure"]}}'
            '}}, '
            '"required": ["action"], '
            '"type": "object"'
        '}}, '
        '"args_format": "Format the arguments as a JSON object."'
    '}}'
'}}\n'
"{tool_list}\n"
"</tools>\n\n"
"For each function call, return a json object with function name and arguments within <tool_call></tool_call> XML tags:\n"
"<tool_call>\n"
'{{"name": <function-name>, "arguments": <args-json-object>}}\n'
"</tool_call>\n\n"
"{important_reminder}\n\n"
"# Response format\n\n"
"Response format for every step:\n"
"1) Action: a short imperative describing what to do in the UI, or specifying which tool to invoke\n"
"2) A single <tool_call>...</tool_call> block containing only the JSON: "
'{{"name": <function-name>, "arguments": <args-json-object>}}.\n\n'
"Rules:\n"
"- Output exactly in the order: Action, <tool_call>.\n"
"- Be brief: one sentence for Action.\n"
"- Do not output anything else outside those parts.\n"
"- If finishing, use action=terminate in the tool call."
)
\end{lstlisting}

\textbf{Message Construction}

\begin{lstlisting}[style=prompt]
# We construct a multimodal message list as follows:
# 1) A system message containing the tool-calling specification. The system prompt is composed of three parts: 
#    (a) the predefined GUI action schema for `computer_use` (always present);
#    (b) optional mcp tool definitions appended from obs["tool_list"] (only when action_space == "mcp" and tools exist);
#    (c) an important-reminder section — either
#    IMPORTANT_WITH_MCP or IMPORTANT_GUI_ONLY — (only when action_space == "mcp", omitted entirely when action_space == "pyautogui").
# 2) For historical context, we keep: steps before the history window [0, t0) as a text-only action history embedded in instruction_prompt ("Step 1: ..., Step 2: ..., ..."); and at most the most recent history_n (= 5) screenshots (image history), each paired with its tool-calling result in a <tool_response> block.
# 3) For each retained past step i in [t0, T): append a user message containing — only for the first retained step i == t0 — instruction_prompt prepended as a text part (task instruction + text-only pre-window action history), followed by a <tool_response> block consisting of the textual tool-calling result tool_calling_results[i] and screenshot[i] as an image_url part; then append an assistant message with the model's response at step i (Action + <tool_call>).
# 4) For the current step T: if T == 0 and history_n > 0 (very first step, no tool has been called yet), append a user message with instruction_prompt and the bare current screenshot as an image_url part (no <tool_response> wrapper); otherwise (T > 0, or T == 0 with history_n == 0), append a user message containing a <tool_response> block with cur_tool_calling_result and the current screenshot as an image_url part, additionally prepending instruction_prompt when history_n == 0 since no prior turn carried it.

# Key variables:
# - T: current step index (0-based), i.e., T = len(actions)
# - history_n: maximum number of past steps retained as image history, fixed at 5
# - t0: start index of the history window, t0 = max(0, T - history_n)
# - screenshots[i]: the processed screenshot at step i, uploaded as an OSS URL
# - tool_calling_results[i]: the textual tool-calling result received at step i, i.e.,
#   obs["exe_result"] after executing step i's action; defaults to "Success" if empty
# - responses[i]: the model's response text at step i (Action + <tool_call>)
# - cur_screenshot: OSS URL of the current screenshot at step T
# - cur_tool_calling_result: tool_calling_results[T], the result received before step T

messages = [
    {"role": "system", "content": [{"type": "text", "text": system_prompt}]}
]

t0 = max(0, T - history_n)

previous_actions_str = "None" if t0 == 0 else "\n".join(
    [f"Step {i+1}: {actions[i]}" for i in range(t0)]
)

instruction_prompt = f"""Please generate the next move according to the UI screenshot, instruction and previous actions.

Instruction: {instruction}

Previous actions:
{previous_actions_str}"""

for i in range(t0, T):
    user_content = []
    if i == t0:
        user_content.append({"type": "text", "text": instruction_prompt})
    user_content += [
        {"type": "text",      "text": "<tool_response>\n"},
        {"type": "text",      "text": tool_calling_results[i]},
        {"type": "image_url", "image_url": {"url": screenshots[i]}},
        {"type": "text",      "text": "\n</tool_response>"},
    ]
    messages.append({"role": "user", "content": user_content})
    messages.append({"role": "assistant", "content": [{"type": "text", "text": responses[i]}]})

if T == 0 and history_n > 0:
    messages.append({
        "role": "user",
        "content": [
            {"type": "text",      "text": instruction_prompt},
            {"type": "image_url", "image_url": {"url": cur_screenshot}},
        ]
    })
else:
    current_user_content = []
    if history_n == 0:
        current_user_content.append({"type": "text", "text": instruction_prompt})
    current_user_content += [
        {"type": "text",      "text": "<tool_response>\n"},
        {"type": "text",      "text": cur_tool_calling_result},
        {"type": "image_url", "image_url": {"url": cur_screenshot}},
        {"type": "text",      "text": "\n</tool_response>"},
    ]
    messages.append({"role": "user", "content": current_user_content})
\end{lstlisting}

\end{tcolorbox}

\end{document}